\documentclass[conference]{IEEEtran}
\usepackage{times}

\usepackage[numbers]{natbib}
\usepackage{multicol}
\usepackage[bookmarks=true]{hyperref}
\usepackage{multirow}

\usepackage{hyperref}
\usepackage{url}
\usepackage{float}

\usepackage{mathtools}
\usepackage{amsfonts}  
\usepackage{graphics}
\usepackage[pdftex]{graphicx}
\usepackage{wrapfig}
\usepackage{caption}
\usepackage{color}
\usepackage{dsfont}
\usepackage[]{mdframed}
\usepackage{algorithm}
\usepackage{algorithmic}
\usepackage{xparse}
\usepackage{amsmath}
\usepackage{bm}
\usepackage{mathtools}
\usepackage{amssymb}
\usepackage{amsthm}
\usepackage{amssymb}

\usepackage{booktabs}
\usepackage{epigraph}

\usepackage{xcolor} 
\usepackage{hyperref} 
\hypersetup{
    colorlinks=true,
}

\usepackage{xargs} 
\usepackage[colorinlistoftodos,prependcaption,textsize=tiny]{todonotes}
\newcommandx{\wrn}[2][1=]{\todo[linecolor=red,backgroundcolor=red!25,bordercolor=red,#1]{#2}}
\newcommandx{\cmt}[2][1=]{\todo[linecolor=blue,backgroundcolor=blue!25,bordercolor=blue,#1]{#2}}

\newcommand{\ba}{\mathbf{a}}
\newcommand{\bv}{\mathbf{v}}
\newcommand{\bu}{\mathbf{u}}
\newcommand{\bo}{\mathbf{o}}
\newcommand{\bq}{\mathbf{q}}
\newcommand{\bI}{\mathbf{I}}
\newcommand{\bA}{\mathbf{A}}
\newcommand{\E}{\mathbb{E}}

\def \ModelSymbol {$\pi_0$}
\def \ModelSymbolBold {$\bm{\pi_0}$}
\def \Robots {7}
\def \Tasks {68}
\IEEEoverridecommandlockouts

\begin{document}
\makeatletter
\let\@oldmaketitle\@maketitle
\renewcommand{\@maketitle}{\@oldmaketitle
  \begin{center}
  \captionsetup{type=figure}
  \includegraphics[width=1.0\textwidth]{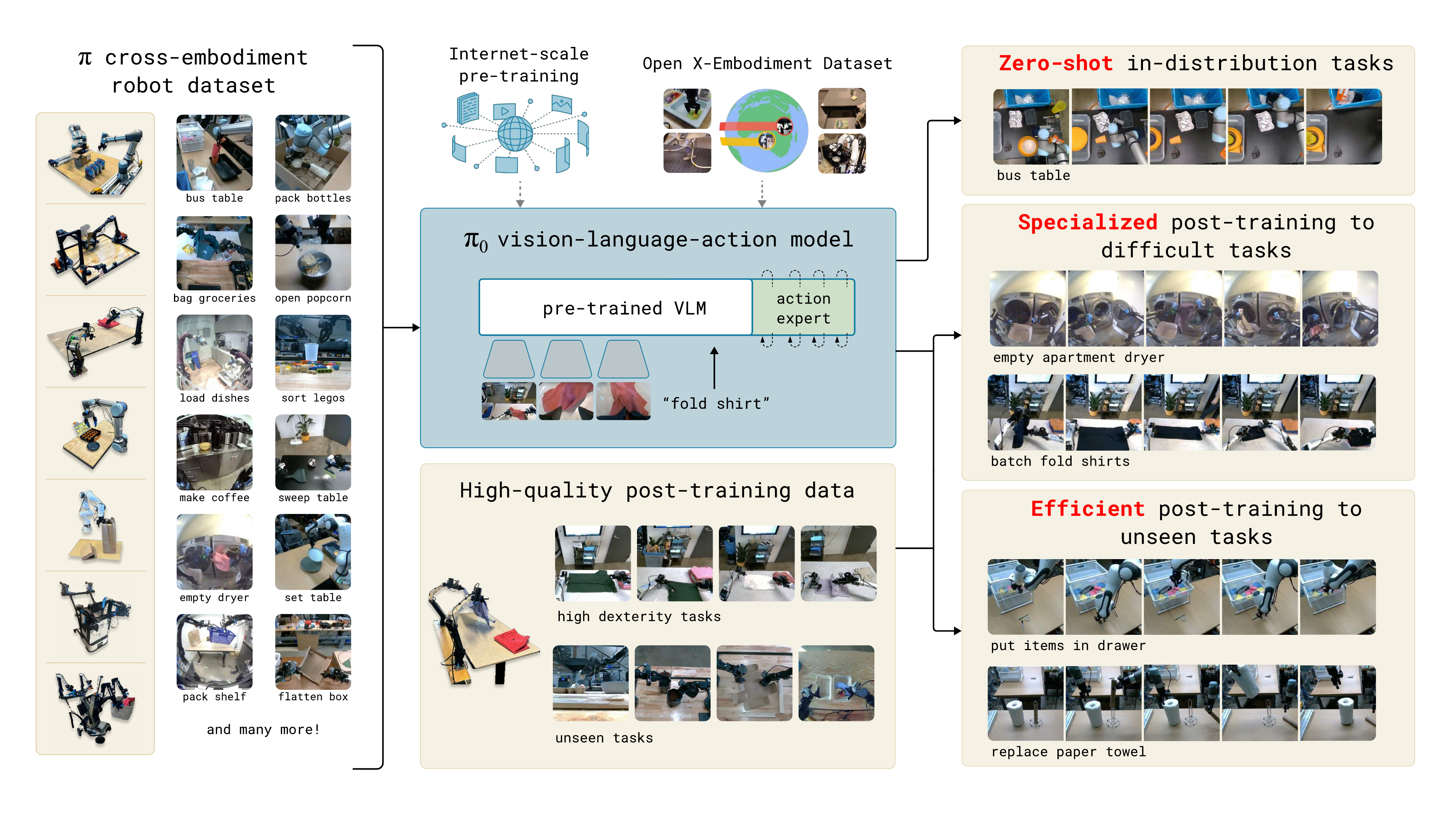}
    \captionof{figure}{Our generalist robot policy uses a pre-trained vision-language model (VLM) backbone, as well as a diverse cross-embodiment dataset with a variety of dexterous manipulation tasks. The model is adapted to robot control by adding a separate \emph{action expert} that produces continuous actions via flow matching, enabling precise and fluent manipulation skills. The model can then be used directly to perform tasks based on a prompt, or fine-tuned on high-quality data to enable complex multi-stage tasks, such as folding multiple articles of laundry or assembling a box.
    } 
    \label{fig:teaser}
    \end{center}
}
\makeatother

\title{
\ModelSymbol: A Vision-Language-Action Flow Model for General Robot Control
}

\pdfinfo{
   /Author (Physical Intelligence)
   /Title  (@title)
   /Subject (Robot Foundation Models)
   /Keywords (Robot Foundation Models)
}


\def\cameraready{0}  

\ifx\cameraready\undefined
    \author{
    Anonymous Submission
    }
\else
    \author{
    \textbf{Physical Intelligence}\\
    Kevin Black,
    Noah Brown,
    Danny Driess,
    Adnan Esmail,
    Michael Equi,
    Chelsea Finn,
    Niccolo Fusai,\\
    Lachy Groom,
    Karol Hausman,
    Brian Ichter,
    Szymon Jakubczak,
    Tim Jones,
    Liyiming Ke,
    Sergey Levine,\\
    Adrian Li-Bell,
    Mohith Mothukuri,
    Suraj Nair,
    Karl Pertsch,
    Lucy Xiaoyang Shi,
    James Tanner,
    Quan Vuong,\\
    Anna Walling,
    Haohuan Wang,
    Ury Zhilinsky
    \thanks{Physical Intelligence, San Francisco, California, USA. Correspondance to: \texttt{research@physicalintelligence.company}}\\
    \vspace{0.05in}
    \url{https://physicalintelligence.company/blog/pi0}
    }
\fi


\maketitle

\begin{abstract}
Robot learning holds tremendous promise to unlock the full potential of flexible, general, and dexterous robot systems, as well as to address some of the deepest questions in artificial intelligence. However, bringing robot learning to the level of generality required for effective real-world systems faces major obstacles in terms of data, generalization, and robustness. In this paper, we discuss how generalist robot policies (i.e., robot foundation models) can address these challenges, and how we can design effective generalist robot policies for complex and highly dexterous tasks. We propose a novel flow matching architecture built on top of a pre-trained vision-language model (VLM) to inherit Internet-scale semantic knowledge. We then discuss how this model can be trained on a large and diverse dataset from multiple dexterous robot platforms, including single-arm robots, dual-arm robots, and mobile manipulators. We evaluate our model in terms of its ability to perform tasks via direct prompting, follow language instructions from people and from a high-level VLM policy, and its ability to acquire new skills via fine-tuning.  Our results cover a wide variety of tasks, such as laundry folding, table cleaning, and assembling boxes.
\end{abstract}

\IEEEpeerreviewmaketitle

\section{Introduction}

\setcounter{figure}{1}

\begin{figure*}[t]
    \centering
    \includegraphics[width=\linewidth]{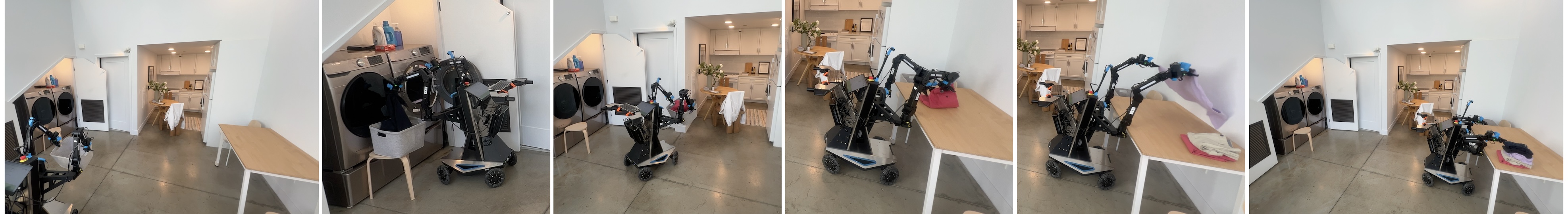}
    \caption{\textbf{\ModelSymbolBold\ controls a mobile manipulator to fold laundry.} Our model is pre-trained on diverse data from \Robots\ distinct robot configurations and \Tasks\ tasks, and can then either be prompted directly or fine-tuned to complex downstream tasks, as in the case of this laundry folding policy, which fetches laundry from a dryer, packs it into a hamper, brings the hamper to a folding table, and then folds each article of clothing.}
    \label{fig:filmstrip}
\end{figure*}

\epigraph{\textit{A human being should be able to change a diaper, plan an invasion, butcher a hog, conn a ship, design a building, write a sonnet, balance accounts, build a wall, set a bone, comfort the dying, take orders, give orders, cooperate, act alone, solve equations, analyze a new problem, pitch manure, program a computer, cook a tasty meal, fight efficiently, die gallantly. Specialization is for insects.}}{Robert A. Heinlein, \textit{Time Enough for Love}}

Artificial intelligence systems come in all shapes and sizes, from highly specialized systems that solve complex problems inaccessible to the human mind, such as predicting the conformation of a protein~\citep{jumper2021highly}, to systems that can produce lifelike high-resolution images or videos based on textual prompts~\citep{rombach2022high}. However, the axis along which human intelligence most outpaces machine intelligence is \emph{versatility}: the ability to solve diverse tasks situated in varied physical environments, while responding intelligently to environmental constraints, language commands, and unexpected perturbations. Perhaps the most tangible progress toward this kind of versatility in AI can be seen in large language- and vision-language models~\citep{achiam2023gpt, team2023gemini}: systems that are pre-trained on large and very diverse corpora of images and text from the web, and then fine-tuned (``aligned'') using more carefully curated datasets meant to induce the desired pattern of behavior and responsiveness. While such models have been shown to exhibit broad instruction-following and problem-solving abilities~\citep{wei2021finetuned, li2022alphacode}, they are not truly \emph{situated} in a physical world the way that people are, and their understanding of physical interaction is based entirely on abstract descriptions. If such methods are to make tangible progress toward AI systems that exhibit the kind of physically situated versatility that people possess, we will need to train them on physically situated data --- that is, data from embodied robot agents.

Flexible and general-purpose models that can be tasked to perform a variety of robot behaviors have tremendous practical ramifications, but they may also offer solutions to some of the toughest challenges facing robot learning today, such as availability of data, generalization, and robustness. In natural language~\citep{achiam2023gpt} and computer vision~\citep{radford2021learning}, general-purpose foundation models that are pre-trained on diverse multi-task data tend to outperform narrowly tailored and specialized solutions. For example, if the goal is to recognize birds in photographs, it is likely more expedient to pre-train on many different image-language associations and then fine-tune or prompt for the bird recognition task, than it is to train on only bird recognition data. Similarly, we may find that for effective specialized robot systems, it is more effective to first pre-train on highly diverse robot data, and then fine-tune or prompt for the desired task. This can resolve the data scarcity challenge, because many more sources of data are available to a generalist model --- including data from other tasks, other robots, or even non-robot sources --- and it may resolve robustness and generalization challenges, because the diverse data exhibits a greater coverage of observations and actions, providing a variety of scenes, corrections, and recovery behaviors that might not be present in more narrow specialized data. Thus, adopting a large-scale pre-training approach to robot learning has the potential to address many of the field's challenges and make practical learning-enabled robots a reality, while at the same time furthering our understanding of the deepest problems in artificial intelligence.

However, developing such generalist robot policies --- i.e., robot foundation models --- involves a number of major challenges. First, any such research must be done at a very large scale, because the full benefits of large-scale pre-training are often not present at smaller scales~\citep{wei2022emergent}. Second, it requires developing the right model architectures that can effectively make use of diverse data sources, while at the same time being able to represent the intricate and subtle behaviors necessary to interact with complex physical scenes. Third, it requires the right training \emph{recipe}. This is perhaps the most important ingredient, as much of the recent progress with large models in NLP and computer vision has relied heavily on delicate strategies for curating pre-training and post-training data~\citep{ouyang2022traininglanguagemodelsfollow}.

In this paper, we present a prototype model and learning framework, which we call \ModelSymbol, that illustrates how each of these three bottlenecks could be tackled. We illustrate our model and system in Figure~\ref{fig:teaser}. To incorporate diverse data sources, we begin by utilizing a pre-trained vision-language model (VLM) to import Internet-scale experience. By basing our model on a VLM, we inherit the general knowledge, semantic reasoning, and problem-solving abilities of language- and vision-language models. 
We then further train our model to incorporate robot actions, turning it into a vision-language-action (VLA) model~\citep{brohan2023rt2visionlanguageactionmodelstransfer}. In order to make it feasible to utilize a variety of diverse robot data sources, we employ \emph{cross-embodiment training}~\citep{collaboration2023open}, where data from many robot types is combined into the same model. These different robot types have different configuration spaces and action representations, including single and dual-arm systems, as well as mobile manipulators. Additionally, in order to make it possible to perform highly dexterous and intricate physical tasks, we use an action chunking architecture~\citep{zhao2023learning} with flow matching (a variant of diffusion) to represent complex continuous action distributions~\citep{lipman2022flow, liu2022rectified}. This enables our model to control robots at frequencies of up to 50 Hz for dexterous tasks such as laundry folding (see Figure~\ref{fig:teaser}). To combine flow matching with VLMs, we use a novel \emph{action expert} that augments the standard VLM with flow-based outputs.

As with language models, the architecture of our model is only part of our method. In order to flexibly and robustly perform complex tasks, we need the right training recipe. Our recipe mirrors the pre-training/post-training separation commonly seen in exascale language- and image-language models~\citep{achiam2023gpt, team2023gemini}, where the model is first pre-trained on a very large and diverse corpus, and then fine-tuned on more narrow and more carefully curated data to induce the desired pattern of behavior --- in our case, dexterity, efficiency, and robustness. Intuitively, training only on high-quality data does not teach the model how to recover from mistakes, since mistakes are rarely seen in such data. Training on only lower-quality pre-training data does not teach the model to act efficiently and robustly. Combining both provides the desired behavior: the model attempts insofar as possible to act in a manner similar to the high-quality data, but still has a repertoire of recoveries and corrections that it can deploy in the case of a mistake.

The contributions of our work consist of a novel generalist robot policy architecture based on VLM pre-training and flow matching, and an empirical investigation of pre-training/post-training recipes for such robot foundation models. We evaluate our model out of the box with language commands, with fine-tuning to downstream tasks, and in combination with a high-level semantic policy that outputs intermediate language commands to perform complex and temporally extended tasks. While our model and system make use of a variety of ideas presented in recent work, the combination of ingredients is novel, and the empirical evaluation demonstrates a level of dexterity and generality that goes significantly beyond previously demonstrated robot foundation models. We evaluate our approach by pre-training on over 10,000 hours of robot data, and fine-tuning to a variety of dexterous tasks, including laundry folding (see Figure~\ref{fig:filmstrip}), clearing a table, putting dishes in a microwave, stacking eggs into a carton, assembling a box, and bagging groceries.

\section{Related Work}
\label{sec:related}

\begin{figure*}[t]
    \centering
    \includegraphics[width=\linewidth]{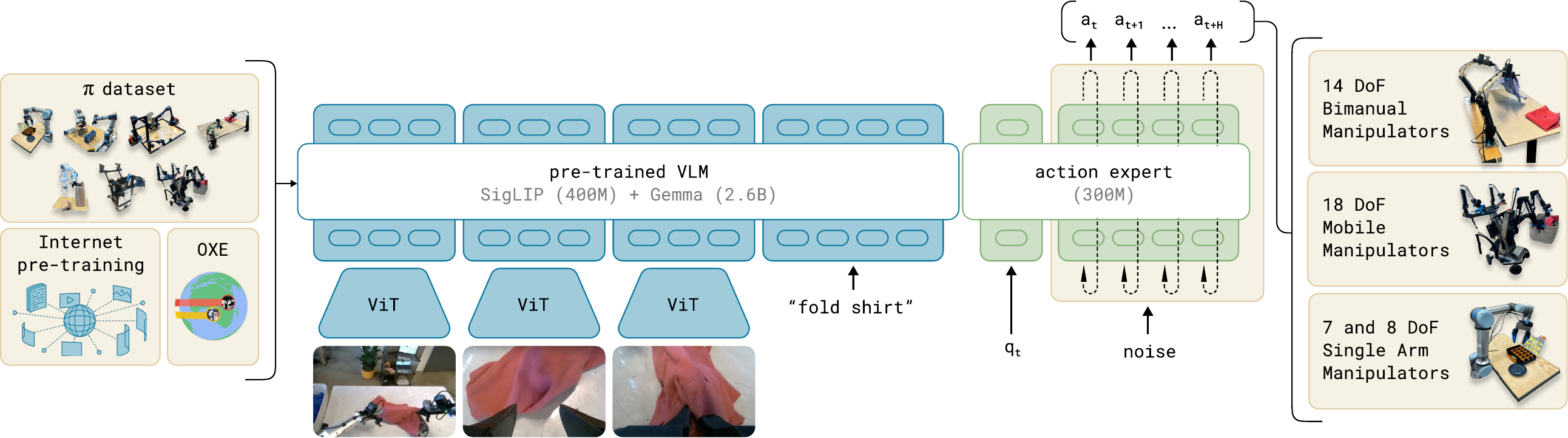}
    \caption{\textbf{Overview of our framework.} We start with a pre-training mixture, which consists of both our own dexterous manipulation datasets and open-source data. We use this mixture to train our flow matching VLA model, which consists of a larger VLM backbone and a smaller \textit{action expert} for processing robot states and actions. The VLM backbone weights are initialized from PaliGemma~\citep{beyer2024paligemma}, providing representations learned from large-scale Internet pre-training. The resulting \ModelSymbol{} model can be used to control multiple robot embodiments with differing action spaces to accomplish a wide variety of tasks.}
    \label{fig:overview}
\end{figure*}

Our work builds on recently proposed methods in large-scale robot learning, as well as multimodal language models. Our work is most closely related to recently proposed vision-language action (VLA) models, which use pre-trained VLMs that are fine-tuned for robot control~\citep{brohan2023rt2visionlanguageactionmodelstransfer, kim2024openvla, wen2024tinyvlafastdataefficientvisionlanguageaction}. Such models employ autoregressive discretization to represent actions in a manner analogous to text tokens. In contrast, our model employs a novel design that fine-tunes a VLM to produce actions via flow matching~\citep{liu2022rectified, lipman2022flow}, a variant of diffusion~\citep{ho2020denoisingdiffusionprobabilisticmodels, sohldickstein2015deepunsupervisedlearningusing}. This allows us to handle high-frequency action chunks~\citep{zhao2023learning} (up to 50 Hz) and highly dexterous tasks, which we show pose a major challenge for prior autoregressive VLAs~\citep{brohan2023rt2visionlanguageactionmodelstransfer}. This resembles a number of recent works on diffusion models for action generation~\citep{chi2023diffusion, zhu2024scalingdiffusionpolicytransformer}. In contrast to these works, our model uses a pre-trained VLM backbone~\citep{beyer2024paligemma}. Our contribution is also fundamentally integrative, focusing on a \emph{framework} for robot foundation models, including not only the model architecture itself but also a pre-training recipe, pre-training and post-training phases, and a range of real-world experiments.

Outside of robot control, many models have been proposed that combine pre-trained language models with diffusion~\citep{rombach2022high, saharia2022photorealistic, esser2024scaling}, including models that specifically hybridize diffusion and autoregressive large language models~\citep{he2024mars, liu2024playground, zhou2024transfusion}. Such models are typically concerned with image generation, but our action generation model builds on a number of previously proposed concepts. Like \citet{zhou2024transfusion}, we train our model via a diffusion-style (flow matching) loss applied on individual sequence elements, in lieu of the standard cross-entropy loss for decoder-only transformers. Like \citet{liu2024playground}, we use a separate set of weights for the tokens corresponding to diffusion. Incorporating these concepts into a VLA model, we introduce what to our knowledge is the first flow matching VLA that produces high-frequency action chunks for dexterous control.

Our work also builds on a rich history of prior works on large-scale robot learning. Early work in this area often utilized self-supervised or autonomous data collection~\citep{levine2016learninghandeyecoordinationrobotic, kalashnikov2018qtoptscalabledeepreinforcement, cabi2020scalingdatadrivenroboticsreward}, providing a tractable data source for simple tasks such as grasping~\citep{gupta2018robotlearninghomesimproving, pinto2015supersizingselfsupervisionlearninggrasp} or pushing~\citep{yu2016more}, but without the complexity of more dexterous behaviors. More recently, a number of high-quality datasets have been collected for robot control that enable broad generalization~\citep{khazatsky2024droid, collaboration2023open, walke2023bridgedata, mandlekar2018roboturkcrowdsourcingplatformrobotic, mandlekar2023mimicgendatagenerationscalable, shafiullah2023bringingrobotshome, ebert2021bridgedataboostinggeneralization, bharadhwaj2023roboagentgeneralizationefficiencyrobot}, but typically for simpler tasks that consist of object relocation and rudimentary furniture manipulation (e.g., drawer opening)~\citep{liu2024ok, etukuru2024robotutilitymodelsgeneral}. More dexterous tasks have been studied at a smaller scale, typically with 10s or 100s of training trajectories~\citep{zhao2023learning}, equivalent to 10 or less hours. Since one of our aims is to study complex and dexterous behaviors, we utilize a much larger dataset, with about 10,000 hours of demonstrations, complemented by the open-source OXE dataset~\citep{collaboration2023open}. To our knowledge, this represents by far the largest robot learning experiment in terms of the amount of robot data. At this scale, we show that a more sophisticated pre-training/post-training recipe is highly effective --- analogously to the recipes used for large language models, a pre-training phase endows our model with a broad base of knowledge, which is then refined in a post-training phase with higher-quality curated data to achieve the desired behavior.

The complexity of the tasks we illustrate goes significantly beyond prior work. While recent work has illustrated a number of more complex and dexterous behaviors, such as tying shoelaces~\citep{zhao2024alohaunleashedsimplerecipe} or cooking shrimp~\citep{fu2024mobile}, we show that our framework can learn very long tasks, sometimes tens of minutes in length, for behaviors that combine both physical dexterity and combinatorial complexity. For example, our laundry folding task requires the robot to manipulate a variety of clothing items that can start in any configuration, and fold multiple items in sequence. Our table bussing task requires discerning the class of novel objects (trash or dishes). We show that a single cross-embodiment model can be used as the base model for these tasks. To our knowledge, our work demonstrates the longest dexterous tasks in the end-to-end robot learning literature.

\section{Overview}
\label{sec:overview}

We provide an outline of our model and training procedure in Figure~\ref{fig:overview}. In our training framework, we first assemble a pre-training mixture consisting of a weighted combination of our own dexterous manipulation datasets (Section~\ref{sec:datadetails}), collected on \Robots\ different robot configurations for \Tasks\ different tasks, and the entire OXE dataset~\citep{collaboration2023open}, which contains data from 22 robots. The pre-training phase (Section~\ref{sec:pre-training}) also uses diverse language labels, combining \emph{task names} and \emph{segment annotations} (fine-grained labels for sub-trajectories, typically about 2 seconds in length). The purpose of the pre-training phase is to train a \emph{base model} that exhibits broad capabilities and generalization, but is not necessarily specialized for high performance on any one task. This base model can follow language commands and perform a variety of tasks at rudimentary proficiency. For complex and dexterous tasks, we then employ a post-training procedure (Section~\ref{sec:pre-training}), which uses high-quality curated data to adapt the model to specific downstream tasks. We study both efficient post-training with small to moderate amounts of data, and high-quality post-training with larger datasets for complex tasks such as laundry folding and mobile manipulation.

Our model, which we describe in Section~\ref{sec:model}, is based on the PaliGemma vision-language model~\citep{beyer2024paligemma}, which we then further train with our data mixture. To turn the base PaliGemma VLM into \ModelSymbol, we add action outputs that use flow matching~\citep{liu2022rectified, lipman2022flow} to generate continuous action distributions. We describe this design in detail in the following section. Note that we use PaliGemma for convenience and because of its comparatively small size (which is useful for real-time control), but our framework is compatible with any base pre-trained VLM.

\section{The \ModelSymbol\ Model}
\label{sec:model}

The \ModelSymbol\ model, illustrated in Figure~\ref{fig:overview}, consists primarily of a language model transformer backbone. Following the standard late fusion VLM recipe~\citep{alayrac2022flamingo, driess2023palm, liu2024visual}, image encoders embed the robot's image observations into the same embedding space as language tokens. We further augment this backbone with robotics-specific inputs and outputs --- namely, proprioceptive state and robot actions. \ModelSymbol\ uses conditional flow matching~\citep{lipman2022flow, liu2022rectified} to model the continuous distribution of actions. Flow matching provides our model with high precision and multimodal modeling capability, making it especially well suited to high-frequency dexterous tasks.
Our architecture is inspired by Transfusion~\citep{zhou2024transfusion}, which trains a single transformer using multiple objectives, with tokens\footnote{In this paper, we use the word ``token'' to refer to an input/output slot along the sequence dimension, whether the slot corresponds to a discrete variable (e.g., a language token) or a continuous variable (e.g., an image patch or a robot action).} corresponding to continuous outputs supervised via a flow matching loss and tokens corresponding to discrete outputs supervised via a cross-entropy loss. Building on Transfusion, we additionally found that using a separate set of weights for the robotics-specific (action and state) tokens led to an improvement in performance. This design is analogous to a mixture of experts~\citep{shazeer2017outrageously, lepikhin2020gshard, du2022glam, fedus2022switch} with two mixture elements, where the first element is used for image and text inputs, and the second is used for robotics-specific inputs and outputs. We refer to the second set of weights as the \emph{action expert}.

Formally, we want to model the data distribution $p(\bA_t | \bo_t)$, where $\bA_t = [\ba_{t}, \ba_{t+1}, ..., \ba_{t+H-1}]$ corresponds to an \emph{action chunk} of future actions (we use $H = 50$ for our tasks), and $\bo_t$ is an observation. The observation consists of multiple RGB images, a language command, and the robot's proprioceptive state, such that $\bo_t = [\bI^1_t, ... , \bI^n_t, \ell_t, \bq_t]$, where $\bI^i_t$ is $i^\text{th}$ image (with 2 or 3 images per robot), $\ell_t$ is a sequence of language tokens, and $\bq_t$ is a vector of joint angles. The images $\bI^i_t$ and state $\bq_t$ are encoded via corresponding encoders and then projected via a linear projection layer into the same embedding space as the language tokens.

For each action $\ba_{t'}$ in the action chunk $\bA_t$, we have a corresponding \emph{action token} that we feed through the action expert. During training, we supervise these action tokens using a conditional flow matching loss~\citep{lipman2022flow, liu2022rectified},
\begin{align*}
    L^\tau(\theta) &= \E_{p(\bA_t | \bo_t), q(\bA_t^\tau | \bA_t)} || \bv_\theta(\bA_t^\tau, \bo_t) - \bu(\bA_t^\tau | \bA_t) ||^2,
\end{align*}
where subscripts denote robot timesteps and superscripts denote flow matching timesteps, with $\tau \in [0, 1]$. Recent work in high-resolution image~\citep{esser2024scaling} and video~\citep{polyak2024movie} synthesis has shown that flow matching can achieve strong empirical performance when combined with a simple linear-Gaussian (or optimal transport) probability path~\citep{lipman2022flow}, given by \mbox{$q(\bA_t^\tau | \bA_t) = \mathcal{N}(\tau \bA_t, (1 - \tau) \mathbf{I})$}. In practice, the network is trained by sampling random noise \mbox{$\epsilon \sim \mathcal{N}(\mathbf{0}, \mathbf{I})$}, computing the ``noisy actions'' \mbox{$\bA_t^\tau = \tau \bA_t + (1 - \tau) \epsilon$}, and then training the network outputs $\bv_\theta(\bA_t^\tau, \bo_t)$ to match the denoising vector field $\bu(\bA_t^\tau | \bA_t) = \bA_t - \epsilon$. The action expert uses a full bidirectional attention mask, so that all action tokens attend to each other. During training, we sample the flow matching timestep $\tau$ from a beta distribution that emphasizes lower (noisier) timesteps. See Appendix~\ref{app:architecture} for more details.

At inference time, we generate actions by integrating the learned vector field from $\tau = 0$ to $\tau = 1$, starting with random noise $\bA_t^0 \sim \mathcal{N}(\mathbf{0}, \mathbf{I})$. We use the forward Euler integration rule:
\begin{align*}
    \bA_{t}^{\tau + \delta} = \bA_{t}^{\tau} + \delta \bv_\theta(\bA_t^\tau, \bo_t),
\end{align*}
where $\delta$ is the integration step size. We use 10 integration steps (corresponding to $\delta = 0.1$) in our experiments. Note that inference can be implemented efficiently by caching the attention keys and values for the prefix $\bo_t$ and only recomputing the suffix corresponding to the action tokens for each integration step. We provide more details regarding the inference procedure, including the inference time for each part of the model, in Appendix~\ref{app:inference}.

While in principle our model can be initialized from scratch or fine-tuned from any VLM backbone, in practice we use PaliGemma~\citep{beyer2024paligemma} as our base model. PaliGemma is an open-source 3 billion parameter VLM that offers a convenient tradeoff between size and performance. We add 300M parameters for the action expert (which is initialized from scratch) for a total of 3.3 billion parameters. We provide a full description of the model architecture in Appendix~\ref{app:architecture}.

\noindent \textbf{Non-VLM baseline model.} In addition to our main VLA model, we also trained a similar baseline model that did not use a VLM initialization for ablation experiments. This model, which we refer to as \ModelSymbol-small, has 470M parameters, does not use VLM initialization, and has a number of small differences that we found to be helpful for training on our data without VLM initialization, which are summarized in Appendix~\ref{app:small_model}. This model is used in our comparisons to evaluate the benefits of incorporating VLM pertaining.

\section{Data Collection and Training Recipe}
\label{sec:recipe}

Broadly capable robot foundation models require not only an expressive and powerful architecture, but also the right dataset and, more importantly, the right training \emph{recipe}. In the same way that LLM training is typically divided into pre-training and post-training phases, we employ a multi-stage training procedure for our model. The goal of the pre-training phase is to expose the model to a diverse range of tasks so that it can acquire broadly applicable and general physical capabilities, while the goal of the post-training phase is to provide the model with the ability to skillfully and fluently execute the desired downstream task. Because of this, the requirements for the pre-training and post-training datasets are distinct: the pre-training dataset should cover as many tasks as possible, and within each of those tasks should cover a diversity of behaviors. The post-training dataset should instead cover behaviors that are conducive to effective task execution, which should exhibit a consistent and fluent strategy. Intuitively, the diverse (but lower quality) pre-training data allows the model to recover from mistakes and handle highly varied situations, which might not otherwise occur in the high-quality post-training data, while the post-training data teaches the model to perform the task well.

\subsection{Pre-training and post-training}
\label{sec:pre-training}

\begin{figure}[h]
    \centering
    \centering
        \includegraphics[width=\linewidth]{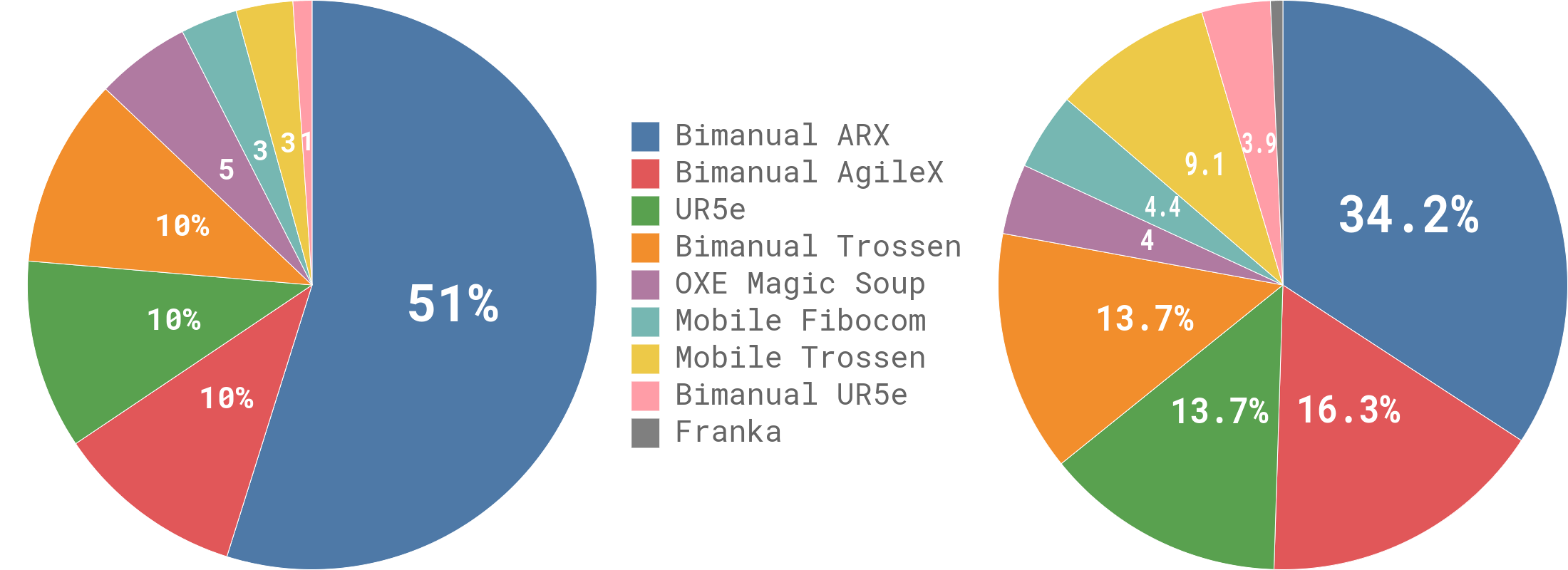} 
    \caption{\textbf{Overview of our dataset}: The pre-training mixture consists of a subset of OXE \cite{collaboration2023open} and the $\pi$ dataset. We use a subset of OXE, which we refer to as OXE Magic Soup \cite{kim2024openvla}. The right figure illustrates the weight of the different datasets in the pre-training mixture. The left figure illustrates their relative sizes as measured by the number of steps.}
    \label{fig:datastats}
\end{figure}

We provide an overview of our pre-training mixture in Figure~\ref{fig:datastats}. Since each training example corresponds to a timestep --- i.e., a tuple $(\bo_t, \bA_t)$, --- we will quantify data in terms of timesteps in this discussion. $9.1\%$ of the training mixture consists of open-source datasets, including OXE~\citep{collaboration2023open}, Bridge v2~\citep{walke2023bridgedata}, and DROID~\citep{khazatsky2024droid}. The robots and tasks in these datasets typically have one or two cameras and use low-frequency control, between 2 and 10 Hz. However, these datasets cover a wide range of objects and environments. To learn dexterous and more complex tasks, we also use 903M timesteps of data from our own datasets, where 106M steps are from single-arm robots and 797M are from dual-arm robots. This data has \Tasks\ tasks, where each task is composed of complex behaviors --- e.g., the ``bussing'' task involves putting a wide range of different dishes, cups, and utensils into a bussing bin, and a wide array of trash items into the garbage. Note that this definition of task is significantly different from prior work, which typically uses any combination of noun and verb (e.g., ``pick up the cup'' vs. ``pick up the plate'') to constitute a distinct task. Therefore, the actual range of behaviors in our dataset is significantly broader than this number of ``tasks'' would imply. We discuss the specific robots and tasks in our dataset in more detail in Section~\ref{sec:datadetails}.

Since the datasets are somewhat imbalanced in size (e.g., the more difficult laundry folding tasks are overrepresented), we weight each task-robot combination by $n^{0.43}$, where $n$ is the number of samples for that combination, such that over-represented combinations are down-weighted. The configuration vector $\bq_t$ and action vectors $\ba_t$ always have the dimensionality of the largest robot in the dataset (18 in our case, to accommodate two 6-DoF arms, 2 grippers, a mobile base, and a vertically actuated torso). For robots with lower-dimensional configuration and action spaces, we zero-pad the configuration and action vectors. For robots with fewer than three images, we also mask out the missing image slots. 

In the post-training phase, we fine-tune our model with a smaller task-specific dataset to specialize it to particular downstream applications. As mentioned previously, our definition of ``task'' is fairly broad --- e.g., the ``bussing'' task requires manipulating a wide range of different objects. Different tasks require very different datasets, with the simplest of the tasks necessitating only 5 hours and the most complex tasks using 100 or more hours of data.

\subsection{Language and high-level policies}
\label{sec:language}

More complex tasks that require semantic reasoning and high-level strategy, such as table bussing, can also benefit from a high-level policy that decomposes high-level tasks (such as ``bus the table'') into more immediate subtasks (such as ``pick up the napkin'' or ``throw the napkin into the trash''). Since our model is trained to process language inputs, we can use a high-level VLM to make these semantic inferences, a method that is analogous to LLM/VLM planning methods such as SayCan~\citep{ahn2022can}. We use such a high-level policy to assist our model with high-level strategy for several of our experimental tasks, as we will discuss in Section~\ref{sec:experiments}.

\subsection{Robot system details}
\label{sec:datadetails}

\begin{figure}[t]
    \centering
    \includegraphics[width=0.95\linewidth]{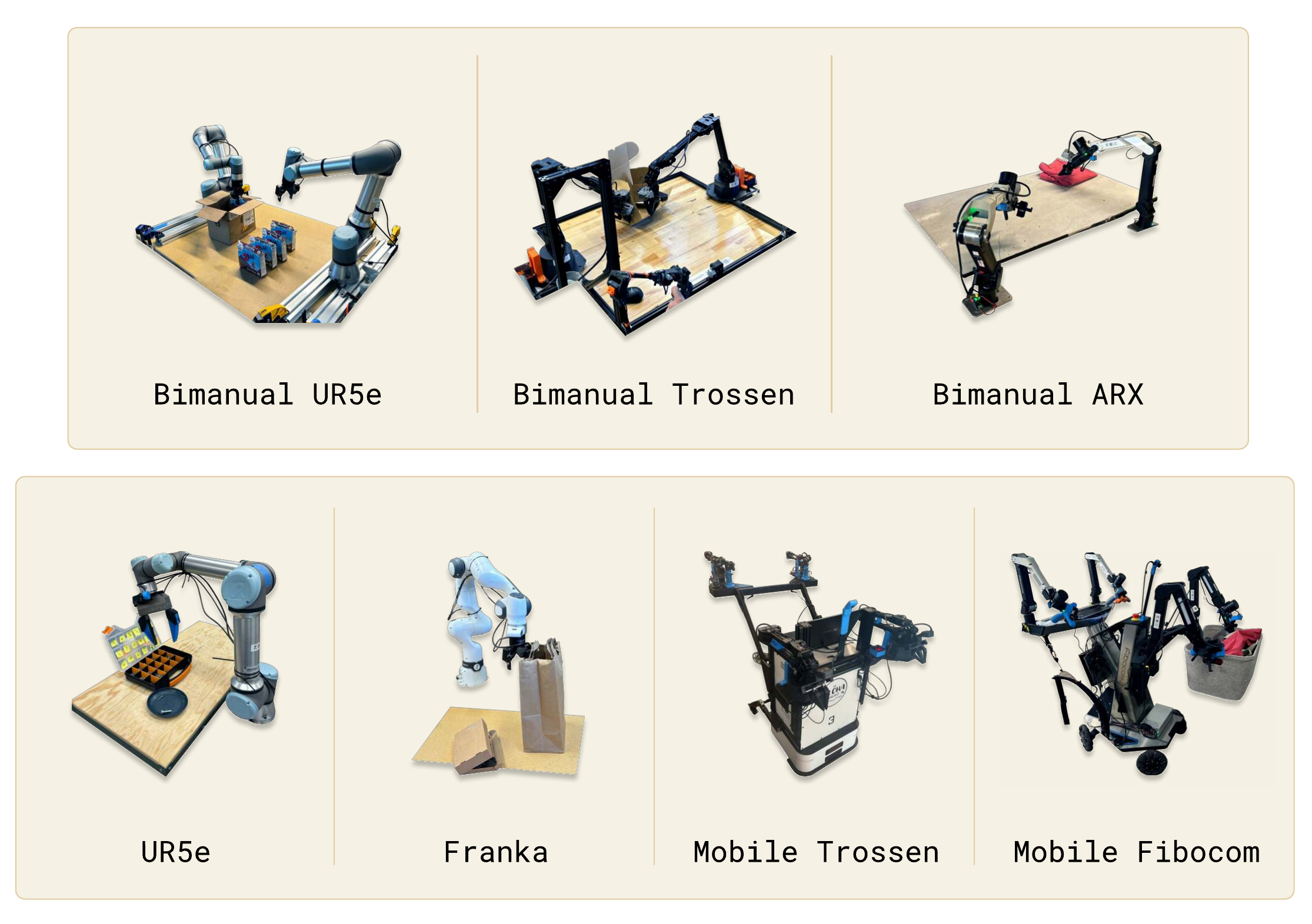}
    \caption{\textbf{The robots used in our experiments.} These include single and dual-arm manipulators with 6-DoF and 7-DoF arms, as well as holonomic and nonholonomic mobile manipulators. \ModelSymbol\ is trained jointly on all of these platforms.}
    \label{fig:robot}
\end{figure}

Our dexterous manipulation datasets include \Robots\ different robot configurations and \Tasks\ tasks. We summarize these platforms in Figure~\ref{fig:robot}, and discuss them below:

\noindent \textbf{UR5e.} An arm with a parallel jaw gripper, with a wrist-mounted and over-the-shoulder camera, for a total of two camera images and a 7-dimensional configuration and action space.

\noindent \textbf{Bimanual UR5e.} Two UR5e setups, for a total of three camera images and a 14-dimensional configuration and action space.

\noindent \textbf{Franka.} The Franka setup has two cameras and an 8-dimensional configuration and action space.

\noindent \textbf{Bimanual Trossen.} This setup has two 6-DoF Trossen ViperX arms in a configuration based on the ALOHA setup~\citep{aldaco2024aloha, zhao2023learning}, with two wrist cameras and a base camera, and a 14-dimensional configuration and action space.

\noindent \textbf{Bimanual ARX \& bimanual AgileX.} This setup uses two 6-DoF arms, and supports either ARX or AgileX arms, with three cameras (two wrist and one base) and a 14-dimensional configuration and action space. This class encompasses two distinct platforms, but we categorize them together because of their similar kinematic properties.

\noindent \textbf{Mobile Trossen \& mobile ARX.} This setup is based on the Mobile ALOHA~\citep{zhao2023learning} platform, with two 6-DoF arms on a mobile base, which are either ARX arms or Trossen ViperX arms. The nonholonomic base adds two action dimensions, for a 14-dimensional configuration and 16-dimensional action space. There are two wrist cameras and a base camera. This class encompasses two distinct platforms, but we categorize them together because of their similar kinematic properties.

\noindent \textbf{Mobile Fibocom.} Two 6-DoF ARX arms on a holonomic base. The base adds three action dimensions (two for translation and one for orientation), for a 14-dimensional configuration and 17-dimensional action space.

We summarize the proportion of our dataset from each robot in Figure~\ref{fig:datastats}.

\section{Experimental Evaluation}
\label{sec:experiments}

Our experimental evaluation consists of out-of-box evaluation experiments that compare our base (pre-trained) model to alternative model designs with direct prompting, as well as detailed fine-tuning experiments that evaluate our model on challenging downstream tasks, comparing it to other methods that have been proposed for dexterous manipulation. We study the following research questions:

\noindent \textbf{How well does \ModelSymbol\ perform after \emph{pre-training} on a variety of tasks that are present in the pre-training data?} We study this question by directly evaluating \ModelSymbol, with comparisons to other robot foundation models.

\noindent \textbf{How well does \ModelSymbol\ follow language commands?} These experiments compare \ModelSymbol\ to \ModelSymbol-small, a smaller version of our model without VLM initialization, to evaluate its performance on following language commands. We evaluate with both human-provided commands and commands specified by a high-level VLM policy, as discussed in Section~\ref{sec:language}.

\noindent \textbf{How does \ModelSymbol\ compare to methods that have been proposed specifically for addressing dexterous manipulation tasks?} These experiments study downstream tasks for which we can either fine-tune our model from the pre-trained initialization, or train it from scratch on task-specific data, comparing to prior methods that were proposed for dexterous manipulation. We aim to evaluate both the benefits of our architecture and our pre-training procedure.

\noindent \textbf{Can \ModelSymbol\ be adapted to complex, multi-stage tasks?} In our final set of experiments, we fine-tune \ModelSymbol\ to a set of particularly complex tasks, including folding laundry and bussing a table. These tasks take between 5 and 20 minutes to complete. Some require guidance from a high-level policy.

\subsection{Evaluating the base model}

\begin{figure}[t]
    \centering
    \includegraphics[width=\linewidth]{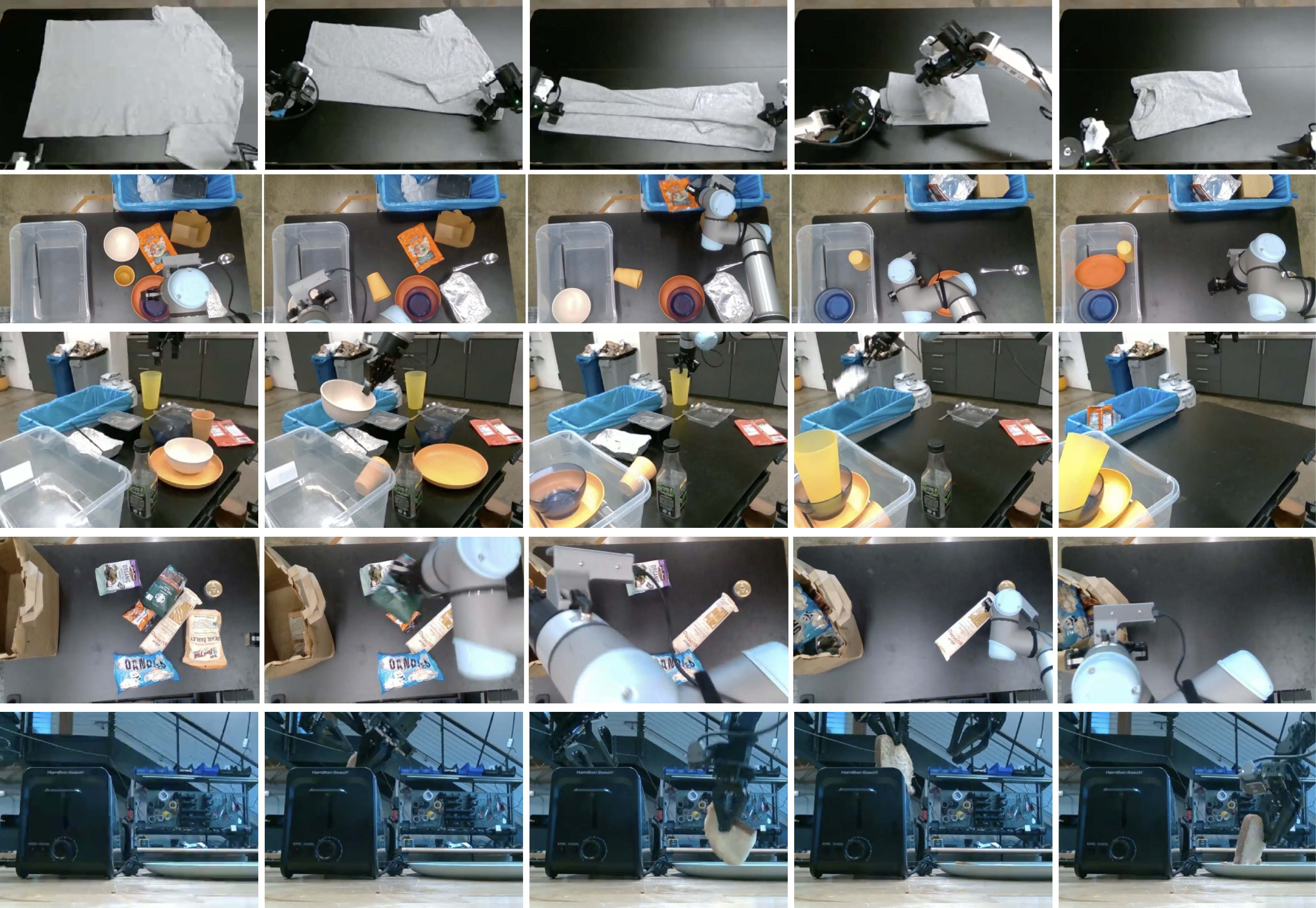}
    \caption{\textbf{Out-of-box evaluation tasks}: To evaluate our base model, we run it after pre-training on five tasks: \textbf{shirt folding}, \textbf{bussing easy}, \textbf{bussing hard}, \textbf{grocery bagging}, and \textbf{toast out of toaster}. The tasks require a combination of dexterous manipulation, multi-stage behaviors, and semantic recognition.}
    \label{fig:pre-train-eval}
\end{figure}

In our first set of experiments, we evaluate the model after pre-training on our full mixture, without any post-training, to evaluate how well our base model can perform a variety of tasks. We compare to other robot foundation models in the literature: both VLAs and smaller models that are trained from scratch on the same pre-training mixture. We evaluate on the following tasks, visualized in Figure~\ref{fig:pre-train-eval}, with each task commanded to the same base model via a language command.

\noindent \textbf{Shirt folding:} the robot must fold a t-shirt, which starts flattened.

\noindent \textbf{Bussing easy:} the robot must clean a table, putting trash in the trash bin and dishes into the dish bin. The score indicates the number of objects that were placed in the correct receptacle.

\noindent \textbf{Bussing hard:} a harder version of the bussing task, with more objects and more challenging configurations, such as utensils intentionally placed on top of trash objects, objects obstructing each other, and some objects that are not in the pre-training dataset.

\noindent \textbf{Grocery bagging:} the robot must bag all grocery items, such as potato chips, marshmallows, and cat food.

\noindent \textbf{Toast out of toaster:} the robot removes toast from a toaster.

Providing comparisons for these experiments is challenging because very few prior models can operate at this scale. We compare to OpenVLA~\citep{kim2024openvla}, a 7B parameter VLA model that was originally trained on the OXE dataset~\citep{collaboration2023open}. We train OpenVLA on our full mixture. This is a very difficult mixture for OpenVLA, which does not support action chunking or high-frequency control. We also compare to Octo~\citep{team2024octo}, a smaller 93M parameter model. While Octo is not a VLA, it does use a diffusion process to generate actions, providing a valuable point of comparison for our flow matching VLA. We also train Octo on the same mixture as our model. Due to time constraints, we were unable to train OpenVLA and Octo for the same number of epochs as our full model. We therefore also compare to a ``compute parity'' version of our model, which is trained for only 160k steps (as opposed to 700k steps for our main model), which is equal to or lower than the number of steps provided to the baselines (160k for OpenVLA, 320k for Octo). We also include a version of the OpenVLA model that we fine-tuned only on the UR5e data, without cross-embodiment training, in the hopes of providing an even stronger baseline on the UR5e tasks. Finally, we include a comparison to the \ModelSymbol-small model described in Section~\ref{sec:model}, which can be viewed as a scaled-down version of our model without VLM pre-training.

\begin{figure}[t]
    \centering
    \includegraphics[width=\linewidth]{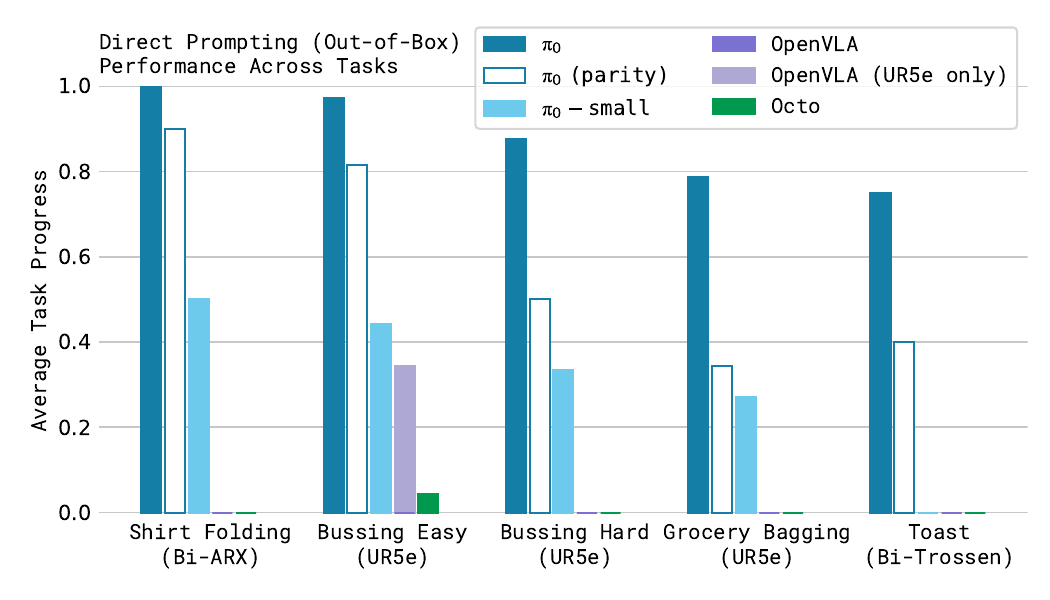}
    \caption{\textbf{Out-of-box evaluation results}: We evaluate \ModelSymbol\ trained for the full 700k steps, a version trained for 160k steps that matches the number of updates for baseline models, \ModelSymbol-small, and three baselines: OpenVLA and Octo trained on all of our data, and OpenVLA trained only on the UR5e tasks (which we found to work better on UR5e tasks). Across all tasks and all comparisons, even the ``parity'' version of our model outperforms all baselines, and the full version of our model achieves the best results by a large margin.}
    \label{fig:pre-train-results}
\end{figure}

The evaluation metric uses a normalized score averaged over 10 episodes per task and method, where an episode receives a score of 1.0 for a full success, and a fractional score for partial success. For example, the score for bussing is the fraction of objects that are correctly placed in the proper receptacle. We describe the scoring rubrics in Appendix~\ref{app:score}. The results, shown in Figure~\ref{fig:pre-train-results}, show that \ModelSymbol\ attains by far the best results across the board on all the out-of-box tasks, with near perfect success rates on shirt folding and the easier bussing tasks, and large improvements over all baselines. The ``parity'' version of \ModelSymbol, which is trained for only 160k steps, still outperforms all the baselines, and even \ModelSymbol-small outperforms OpenVLA and Octo. OpenVLA struggles on these tasks because its autoregressive discretization architecture does not support action chunks. The UR5e-only OpenVLA model performs better, but is still far below the performance of \ModelSymbol. Octo does support action chunks, but has a comparatively limited representational capacity. This comparison illustrates the importance of combining large, expressive architectures with the ability to model complex distributions via flow matching or diffusion. Additionally, the comparison to \ModelSymbol-small illustrates the importance of incorporating VLM pre-training. Unfortunately, it is hard to make this last comparison fair: \ModelSymbol-small uses fewer parameters, but larger models are difficult to use without pre-training. Overall, these experiments show that \ModelSymbol\ provides a powerful pre-trained model with the ability to effectively perform a variety of tasks with a variety of robots, with much better performance than prior models.

\subsection{Following language commands}

In the next set of experiments, we fine-tune the base \ModelSymbol\ model to follow language commands in a set of evaluation domains. We compare this fine-tuned \ModelSymbol\ model with the \ModelSymbol-small model described in Section~\ref{sec:model}, which we found to be the strongest baseline in the previous section. Recall that \ModelSymbol-small does \emph{not} use a VLM initialization. This experiment therefore aims to measure how much VLM pre-training boosts our model's ability to follow language instructions. Note that \ModelSymbol-small is also a significantly smaller model --- unfortunately, it is difficult to remove this confounder, because VLM initialization serves both to make it practical to train a much larger model without overfitting, and to improve language instruction following. We nonetheless hope that this experiment sheds light on the language capabilities of \ModelSymbol. The language instructions for each task consist of objects to pick up and locations to place those objects, with language-labeled segments that are about 2 seconds in length. Each full task consists of numerous such segments. The tasks in this evaluation consist of:

\noindent \textbf{Bussing:} the robot must clean a table, placing dishes and cutlery in a bin, and trash into a trash bin.

\noindent \textbf{Table setting:} the robot must take out items from a bin to set a table, including a place mat, dishes, silverware, napkin, and cups, and adjust them according to language instructions.

\noindent \textbf{Grocery bagging:} the robot must pack grocery items, such as bags of coffee beans, barley, marshmallow, seaweed, almonds, spaghetti, and cans into a bag.

\begin{figure}
    \includegraphics[width=\linewidth]{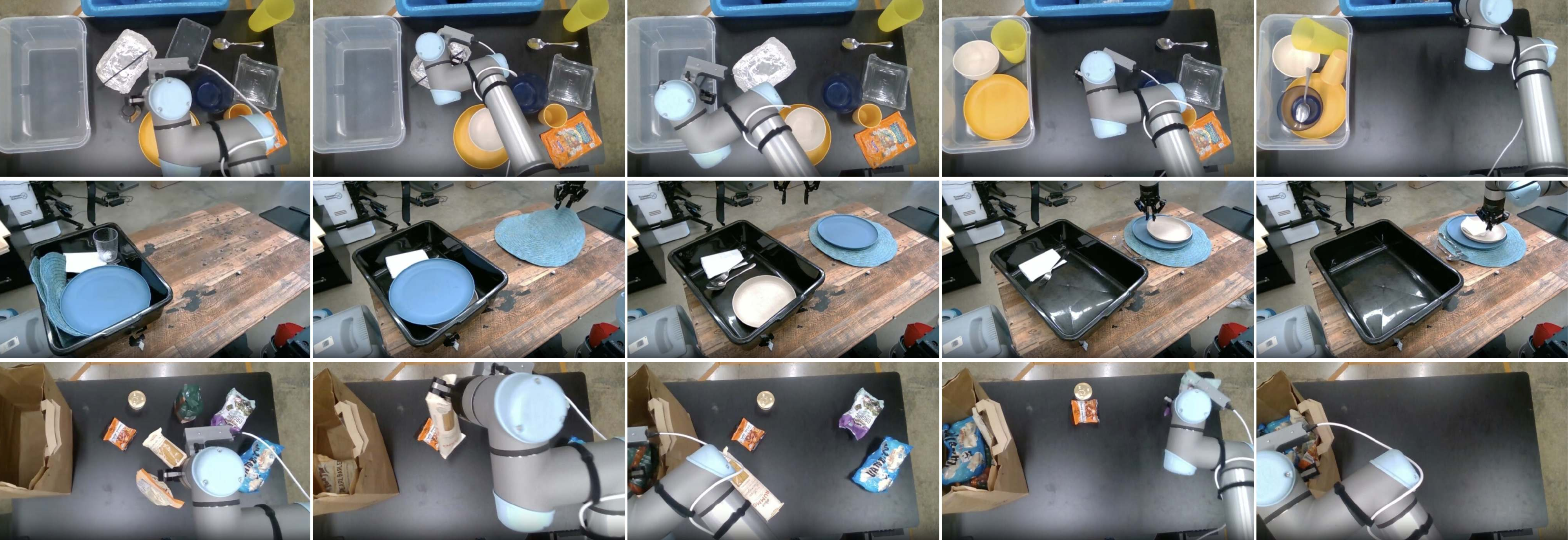}
    \caption{\textbf{The tasks in our language evaluation.} We evaluate our model on 3 different language-conditioned tasks, each of which requires following a sequence of intermediate language commands. The tasks involve bussing a table (top) to put dishes in a bin and garbage in a trash bin, setting a table (middle) by taking items out of a bin, and packing a shopping bag (bottom).}
    \label{fig:eval-language}
\end{figure}

In Figure~\ref{fig:eval-language}, we show the language-conditioned tasks in our evaluation and present the evaluation results. We evaluate five different conditions. \ModelSymbol-flat (and \ModelSymbol-small-flat) corresponds to directly command the model with the task description (e.g., ``bag the groceries''), without intermediate language commands. \ModelSymbol-human (and \ModelSymbol-small-human) provides intermediate step commands (e.g., which object to pick and where to place it) from an expert human user. These conditions evaluate each model's ability to follow more detailed language commands: while these intermediate commands provide considerable information for how to perform the task, the model must be able to understand and follow those commands to benefit from them. Finally, \ModelSymbol-HL evaluates \ModelSymbol\ with high-level commands provided by a high-level VLM, as discussed in Section~\ref{sec:language}. This condition is also autonomous, without any human expert.

\begin{figure}
    \includegraphics[width=\linewidth]{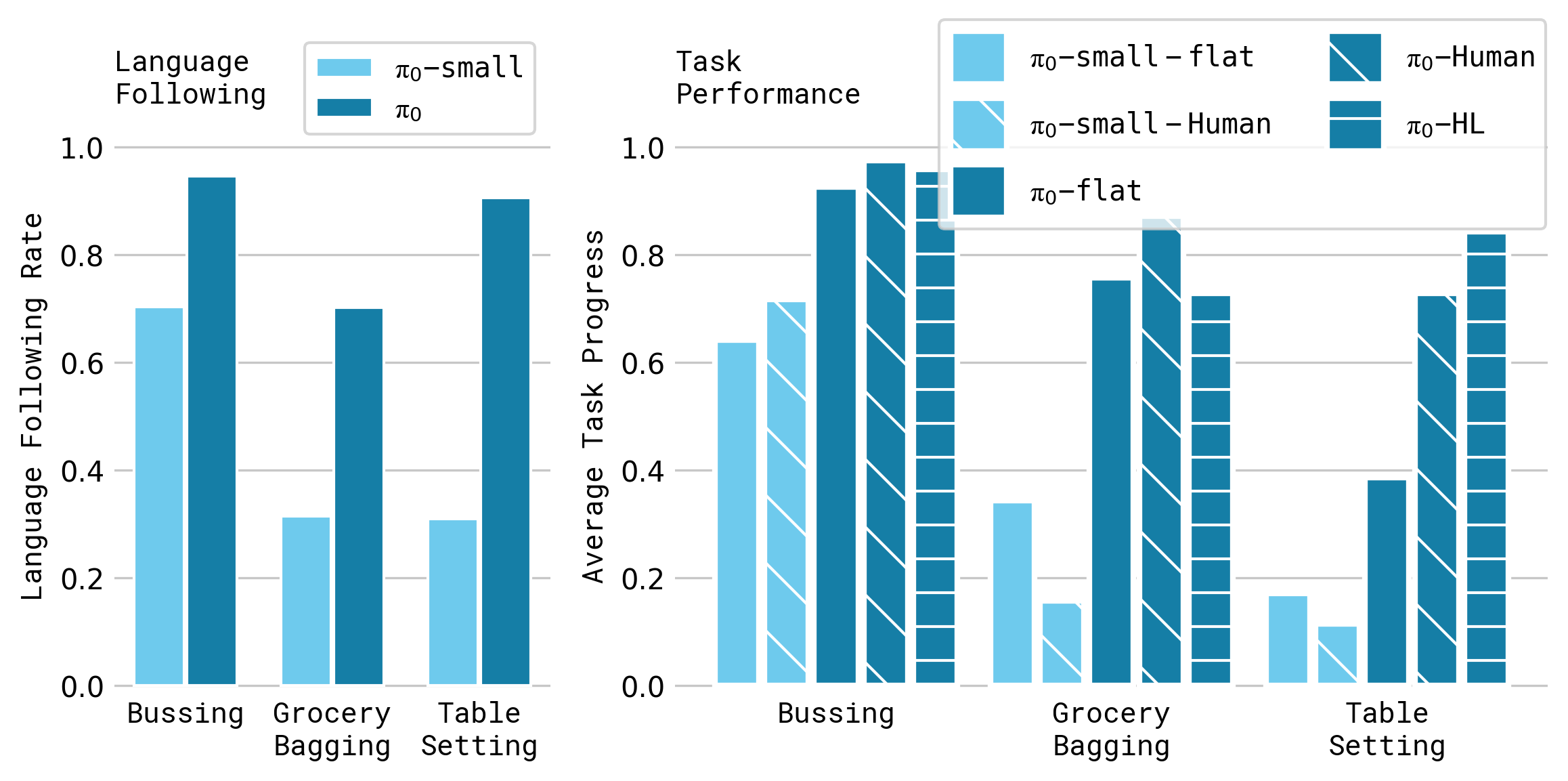}
    \caption{\textbf{Language evaluation.} We compare ``flat'' versions of our policies, $\mathtt{-flat}$, which receive only the overall task command (e.g., ``bag the groceries'') with a method that receives intermediate commands from a human expert, $\mathtt{-human}$, or a high-level VLM policy, $\mathtt{-HL}$. We also compare our model to a small non-VLM variant under the ``expert'' condition, \ModelSymbol~and \ModelSymbol-small, in terms of language following accuracy. The results show a significant improvement with \ModelSymbol\ from intermediate language commands provided by a human expert and to a lesser degree by an autonomous high-level policy. Notably, due to \ModelSymbol-small's limited language following ability, overall it does not gain with the addition of a high-level expert.}
    \label{fig:eval-language-results}
\end{figure}

The results in Figure~\ref{fig:eval-language-results}, averaging over 10 trials per task, show that the language following accuracy of \ModelSymbol\ is significantly better than that of \ModelSymbol-small. This suggests a significant improvement from the larger pre-trained VLM initialization. This capability translates to an improvement in performance with expert human guidance (\ModelSymbol-human) and with high-level model guidance (\ModelSymbol-HL). The results indicate that \ModelSymbol's language following ability directly translates into better autonomous performance on complex tasks with high-level guidance.

\subsection{Learning new dexterous tasks}

\begin{figure}[t]
    \centering
    \includegraphics[width=\linewidth]{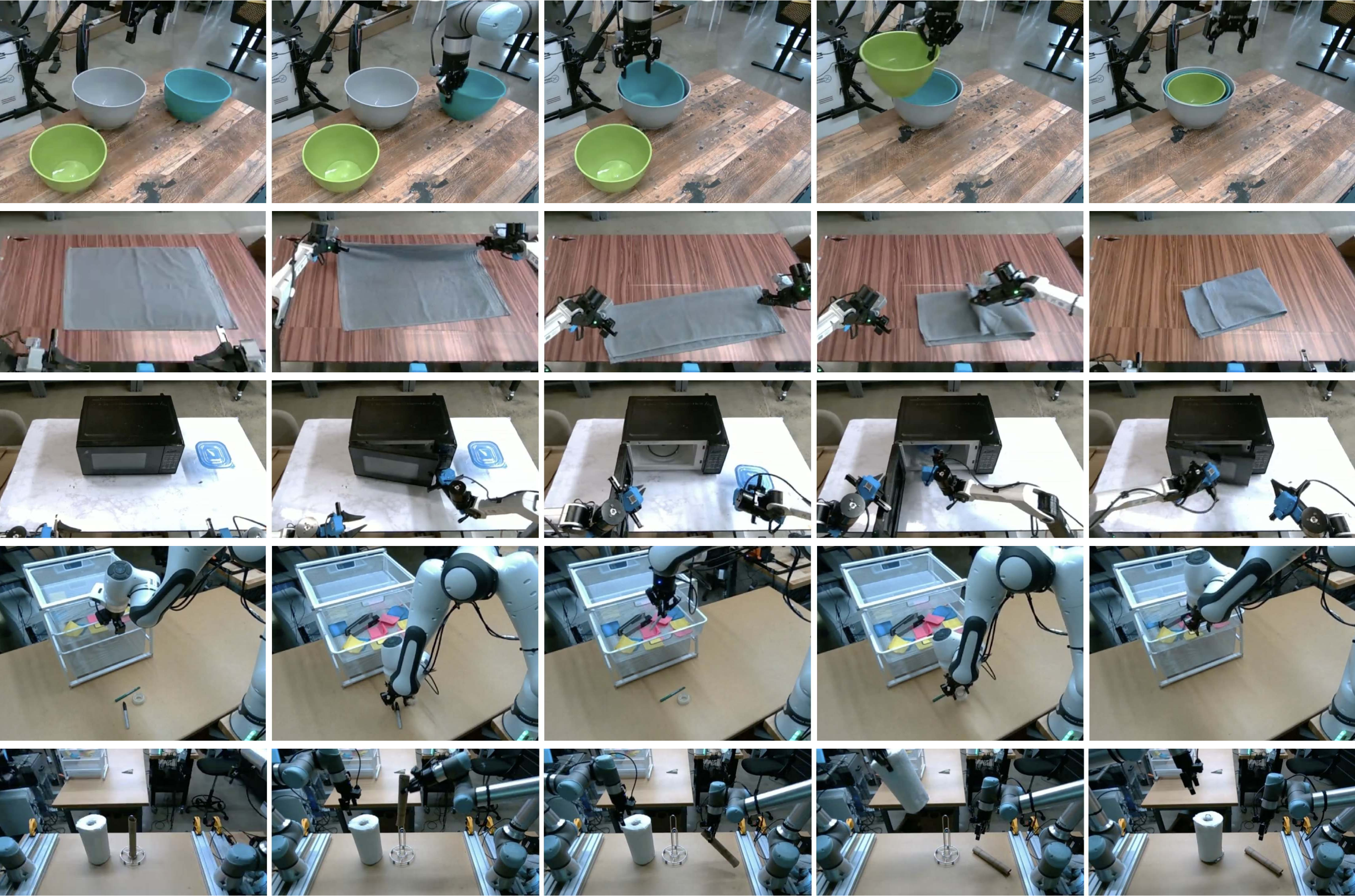}
    \caption{\textbf{Fine-tuning evaluation tasks}: We fine-tune our model to a variety of downstream tasks that are distinct from the tasks seen in pre-training. Our tasks represent a range of similarity from the pre-training tasks, with tasks that are most similar to pre-training (\textbf{stack bowls} and \textbf{towel folding}), a task that introduces an unseen new element (a microwave), and tasks that require new motions and new object types (\textbf{Franka items in drawer} and \textbf{paper towel replacement}).}
    \label{fig:fine-tuning-eval}
\end{figure}

In the next set of experiments, we evaluate our model on new tasks that differ significantly from the pre-training data, requiring entirely new behaviors. For these evaluations, we fine-tune the model using various amounts of data for each new task. While each task is new, we partition the tasks into ``tiers'' depending on how much they differ from tasks in the pre-training data. The tasks, shown in Figure~\ref{fig:fine-tuning-eval}, are:

\noindent \textbf{UR5e stack bowls.} This task requires stacking bowls, with four bowls of different sizes. Since this task requires grasping and moving dishes like the bussing task in the pre-training data, we place it in the ``easy'' tier. The training data contains a variety of bowls, and the evaluations use a mix of seen and unseen bowls.

\noindent \textbf{Towel folding.} This task requires folding a towel. Since this is similar to shirt folding, which is present in pre-training, we place it in the ``easy'' tier.

\noindent \textbf{Tupperware in microwave.} This task requires opening a microwave, putting a plastic container inside it, and closing it. The containers come in different shapes and colors, and the evaluations use a mix of seen and unseen containers. The container manipulation resembles pre-training data, but the microwave is not found in pre-training.

\noindent \textbf{Paper towel replacement.} This task requires removing an old cardboard paper towel tube from a holder and replacing it with a fresh paper towel roll. Because no such items are found in pre-training, we consider this ``hard.''

\noindent \textbf{Franka items in drawer.} This task requires opening a drawer, packing items into a drawer, and closing it. Because there is no similar task with the Franka robot in pre-training, we consider this ``hard.''

\begin{figure*}[t]
    \centering
    \includegraphics[width=0.9\linewidth]{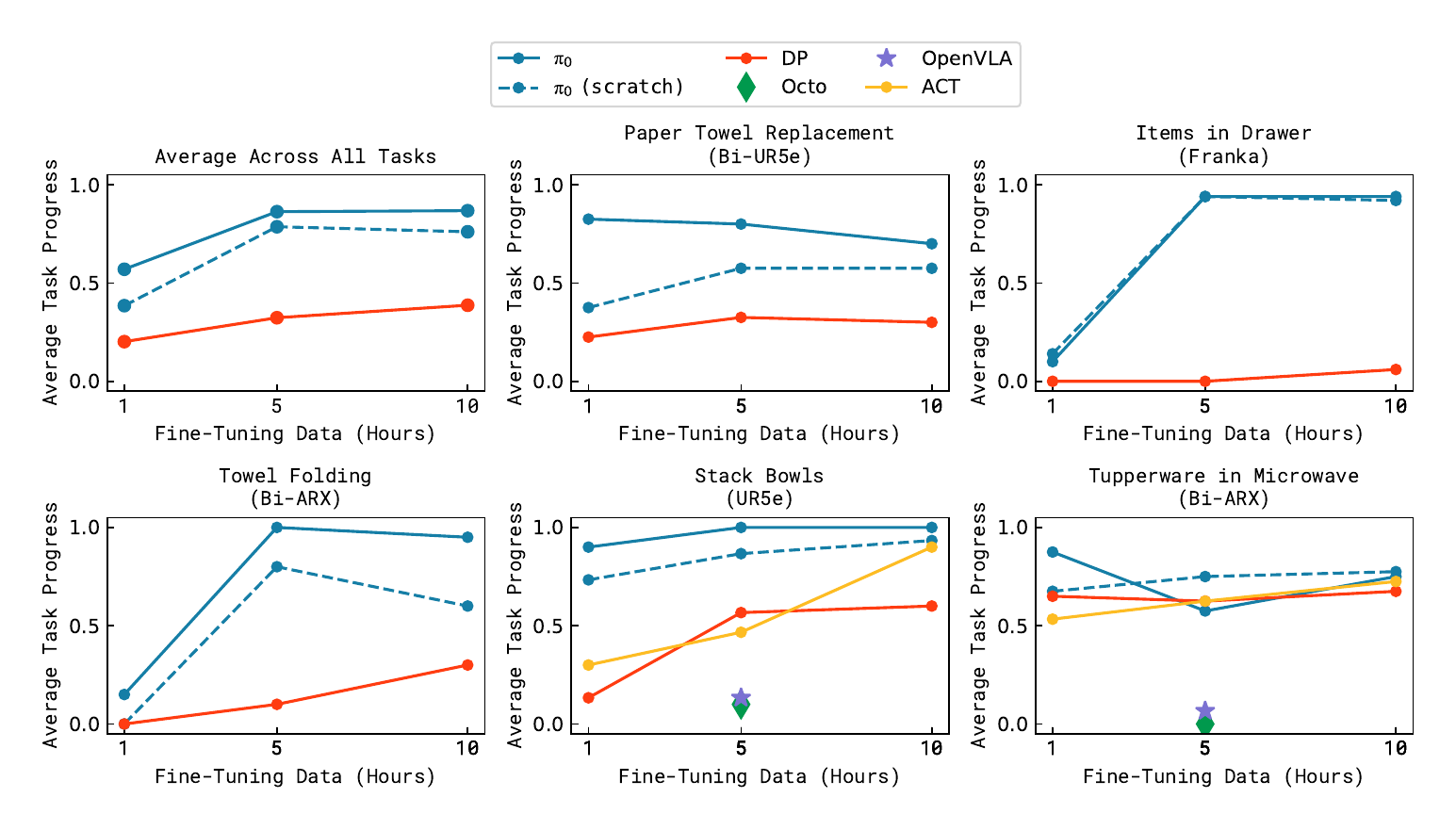}
    \caption{\textbf{Fine-tuning with varying amounts of data.} \ModelSymbol\ can learn some easier tasks even with smaller amounts of data, and the pre-trained model often attains a larger improvement over the model trained from scratch.}
    \label{fig:fine-tuning-sweep}
\end{figure*}

We compare our model after fine-tuning both to OpenVLA~\citep{kim2024openvla} and Octo~\citep{team2024octo}, which also employ a pre-training and fine-tuning recipe. Since our aim is to evaluate the specific models (rather than the architectures), we use the publicly available pre-trained checkpoints for these models, which are trained on OXE~\citep{collaboration2023open}, and then fine-tune them to each task. We also compare to ACT~\citep{zhao2023learning} and Diffusion Policy~\citep{chi2023diffusion}, which are designed specifically for learning dexterous tasks from smaller datasets. ACT and Diffusion Policy are trained \emph{only} on the fine-tuning datasets, which are of similar size to the individual datasets used in the ACT and Diffusion Policy experiments~\citep{chi2023diffusion, zhao2023learning}. We evaluate \ModelSymbol\ by fine-tuning from our pre-trained base model, as well as by training from scratch. This comparison is meant to evaluate the individual benefits of the \ModelSymbol\ architecture and our pre-training procedure. We hypothesize that the \ModelSymbol\ architecture with VLM initialization should already provide a stronger starting point for the individual tasks, while the pre-training procedure should further improve its performance, especially with smaller fine-tuning datasets.

Figure~\ref{fig:fine-tuning-sweep} shows the performance across all of the tasks for a variety of methods, averaging over 10 trials per task, with different amounts of fine-tuning data on each task. We include all of the baselines on the \textbf{stack bowls} and \textbf{Tupperware in microwave} tasks. Since OpenVLA and Octo attain significantly worse performance, we only run these for one of the dataset sizes, due to the time cost of evaluating so many models in the real world. The results show that \ModelSymbol\ generally outperforms other methods. Interestingly, the strongest prior models are the ones that are trained entirely from scratch on the target tasks, suggesting that leveraging pre-training in these domains presents a major challenge for prior approaches. While the 5-hour policy for \ModelSymbol\ on the Tupperware task performs similarly to the baselines, the 1-hour version is significantly better. As expected, pre-training leads to larger improvement for tasks that are more similar to the pre-training data, though the pre-trained model is frequently better than the non-pre-trained model, sometimes by as much as 2x.

\subsection{Mastering complex multi-stage tasks}

\begin{figure}[t]
    \centering
    \includegraphics[width=\linewidth]{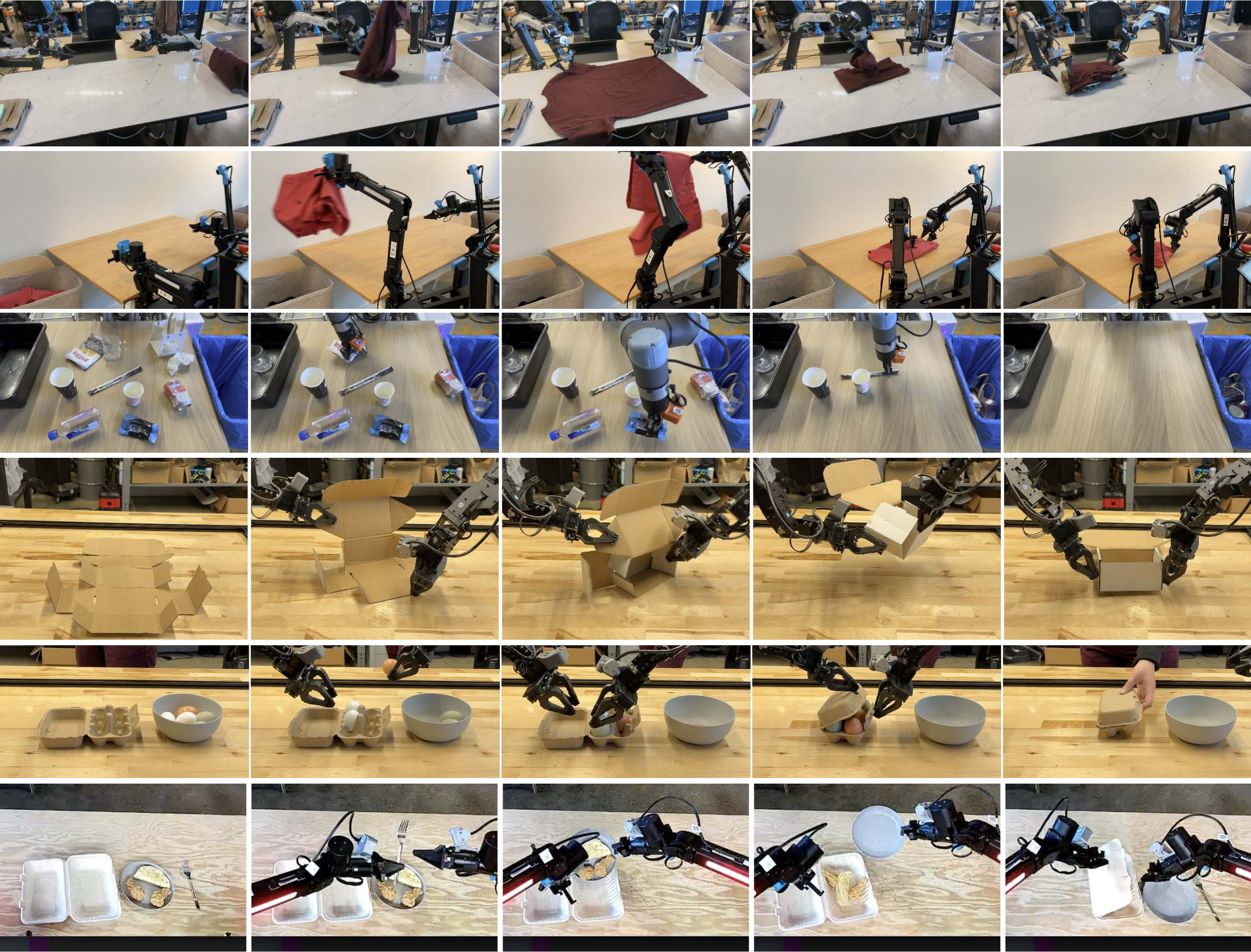}
    \caption{\textbf{We evaluate a range of complex and temporally extended tasks.} This includes: folding laundry from a bin with a stationary (a) or mobile (b) robot, bussing a real lunch table (c), assembling a box (d), packing eggs into a carton (e), and packing food into a to-go box (f). These tasks require combining dozens of individual behaviors, such as grasping, stacking, folding, and flattening, generalization to a huge variety of object configurations, and complex physical properties, such as deformable objects or flexible cardboard.}
    \label{fig:post-training}
\end{figure}

In our final set of experiments, we tackle a range of challenging multi-stage tasks via a combination of fine-tuning and language. For some of these tasks, data is present in pre-training, but fine-tuning is required to attain mastery. For some, no data is present in pre-training. The tasks in this evaluation, shown in Figure~\ref{fig:post-training}, are:

\begin{figure}[t]
    \centering
    \includegraphics[width=\linewidth]{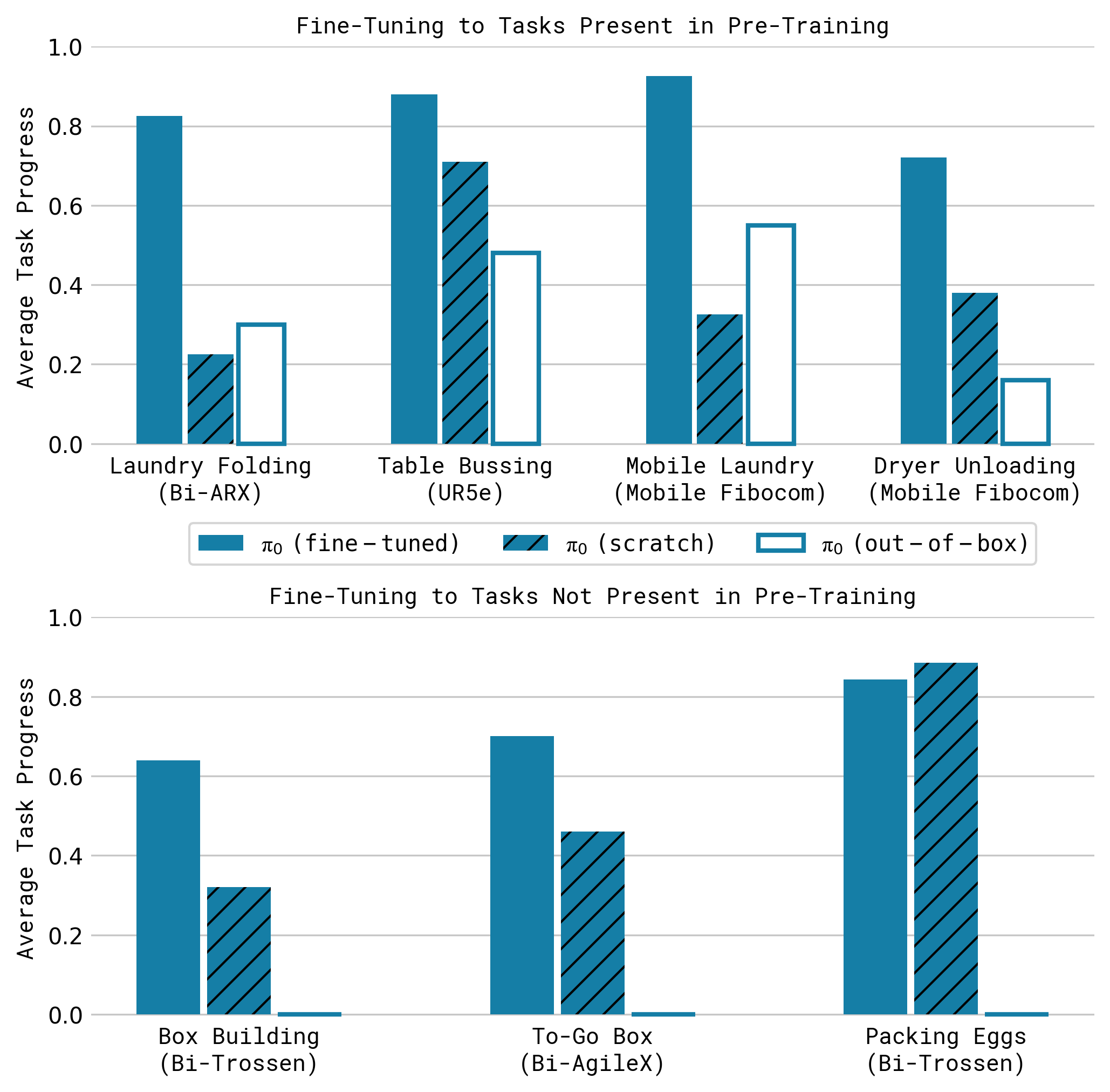}
    \caption{\textbf{Post-training results on complex tasks} in terms of average scores over 10 trials. The full pre-trained \ModelSymbol\ model attains more than 50\% of the maximum score across all of the tasks, and typically outperforms the ablations, with especially significant improvements on the hardest tasks.}
    \label{fig:post-training-results}
\end{figure}

\noindent \textbf{Laundry folding:} This task requires a static (non-mobile) bimanual system to fold articles of clothing. The clothing items start in a randomized crumpled state in a bin, and the goal is to take out the item, fold it, and place it on top of a stack of previously folded items. The randomized initial configuration of the crumpled laundry presents a major challenge, since the policy needs to generalize to any configuration. This task is present in pre-training.

\noindent \textbf{Mobile laundry:} Here, the Fibocom mobile robot in Figure~\ref{fig:robot} has to fold laundry, facing many of the same challenges while controlling orientation and translation. This task is present in pre-training.

\noindent \textbf{Dryer unloading:} Here, the Fibocom mobile robot has to take laundry out of a dryer and place it into a hamper. This task is present in pre-training.

\noindent \textbf{Table bussing:} This task requires bussing a table with a diverse array of novel objects in a clutter scene, presenting a much greater challenge than the benchmark in our out-of-box evaluation: the policy must generalize to unseen objects of varying shapes and sizes, and perform complex dexterous motions, such as twisting the gripper to pick up large plates and carefully grasping thin, delicate items such as glasses. The robot must handle dense clutter and intelligently sequence various behaviors --- for example, to clean off a plate with trash, it must first pick up the plate, then shake its contents into the garbage, and then place the plate in the bin. This task is not present in pre-training.

\noindent \textbf{Box building:} The robot has to assemble a cardboard box that starts in a flattened state. This task presents a number of major challenges: the box needs to bent in the right way, and the robot needs to hold down parts of the box while folding others, utilizing both arms and even the surface of the table to brace during folding motions. The robot might need to retry some folds, requiring a reactive and intelligent strategy. This task is not present in pre-training.

\noindent \textbf{To-go box:} This task requires moving several food items from a plate into a to-go box, requiring packing the items into the box so that they do not stick out, and then closing the box with both arms. This task is not present in pre-training.

\noindent \textbf{Packing eggs:} The robot needs to take six eggs out of a bowl and pack them into an egg carton, and then close the carton. The eggs need to be grasped in a manner appropriate to their pose inside the bowl, and then placed into open slots in the carton. This presents challenges due to the egg shape, slipperiness, and the need for careful placement. Closing the box requires the use of both arms. This task is not present in pre-training.

The results, showing average scores per task over 10 trials, are presented in Figure~\ref{fig:post-training-results}. The scoring rubrics are in Appendix~\ref{app:score}. A score of 1.0 represents a perfect execution, while partial scores correspond to partially completed tasks (e.g., 0.5 indicates that half the objects were bussed correctly). These tasks are very difficult, and we were not able to solve them with other methods. We therefore use these tasks to compare to ablations of our approach, evaluating \ModelSymbol\ after pre-training and fine-tuning, out of the box after pre-training only (``out-of-box''), and training on the fine-tuning data without any pre-training (``scratch''). The results show that \ModelSymbol\ can solve many of these tasks, with our full pre-training and fine-tuning recipe performing best across the board. Note that many of these more difficult tasks show a very large improvement from using the pre-trained model, indicating that pre-training is especially useful with harder tasks. The absolute performance of \ModelSymbol\ varies across the tasks, likely due to differences in task difficulty and the degree to which the tasks are represented in pre-training. We recommend that readers watch the task videos on the \href{https://physicalintelligence.company/blog/pi0}{accompanying website} for a more complete impression of these tasks and their complexity. We believe that this level of autonomous performance on such challenging tasks represents a new state of the art in dexterous robot manipulation with learned policies.

\section{Discussion, Limitations, and Future Work}

We presented a framework for training a robot foundation model, which we refer to as \ModelSymbol, that consists of pre-training on highly diverse data, followed by either out-of-box evaluation or fine-tuning to complex downstream tasks. Our empirical evaluation studies tasks that combine dexterity, generalization, and temporally extended multi-stage behaviors. Our model incorporates Internet-scale vision-language model (VLM) pre-training with flow matching for representing complex high-frequency action chunks. Our pre-training mixture consists of 10,000 hours of dexterous manipulation data from \Robots\ different robot configurations and \Tasks\ tasks, in addition to large amounts of previously collected robot manipulation data from OXE~\citep{collaboration2023open}, DROID~\citep{khazatsky2024droid}, and Bridge~\citep{walke2023bridgedata}. To our knowledge, this represents the largest pre-training mixture ever used for a robot manipulation model. Our fine-tuning experiments include over 20 tasks, where we show that our model outperforms a variety of baselines, including prior VLA models~\citep{kim2024openvla} and models designed specifically for dexterous manipulation~\citep{zhao2023learning, chi2023diffusion}. We also examine how our post-training recipe can enable highly complex tasks, such as folding multiple articles of clothing from arbitrary initial configurations or assembling boxes.

Our framework broadly resembles the training procedures employed for large language models, which typically consist of pre-training a base model on very large datasets scraped from the web, followed by a post-training procedure that aims to ``align'' the model to enable it to follow instructions and perform user commands. It is generally recognized that most of the ``knowledge'' in such models is acquired in the pre-training phase, while the post-training phase serves to tell the model how it should leverage that knowledge to fulfill user commands. Our experiments imply that an analogous phenomenon might take place with robot foundation models, where pre-trained models have some zero-shot capabilities, but complex tasks like laundry following require fine-tuning with high-quality data. Training on only this high-quality data results in a brittle model that does not reliably recover from mistakes, while running the pre-trained model in zero shot does not always exhibit the fluent strategies demonstrated in the post-training data.

We hope that our results will serve as a stepping stone toward general and broadly applicable robot foundation models. Our experiments suggest that such models may soon be a reality, but there are a number of limitations and ample room for future work. First, our experiments do not yet provide a comprehensive understanding of how the pre-training datasets should be composed: we combined all data available to us, but understanding what type of data is more helpful to add and how it should be weighted remains an open problem. Not all tasks in our evaluation work reliably, and it remains unclear how to predict how much and what kind of data is needed to attain near-perfect performance. Finally, it remains to be seen how much positive transfer there is in combining highly diverse data, particularly from different tasks and different robots: although our results suggest that universal pre-trained robot foundation models might become a reality, it is left for future work to understand whether this universality extends to much more distinct domains, such as autonomous driving, navigation, and legged locomotion.

\section*{Acknowledgements}

We thank Laura Smith and Dibya Ghosh for feedback on the paper and assistance with figures and videos, Philip Clark, Kelly Sims, and Saunaz Moradi for feedback on writing, and Evan Pokrandt, Joakim Keussen, Dan Philibin, Eitan Penner, Adam Lisagor, and Greg Miller for help with illustrations, design, and videos. We also thank Lili Yu for helpful technical discussion. We are tremendously grateful to all of the robot operators for tirelessly collecting robot manipulation data. For a full contribution statement, see Appendix~\ref{app:contributions}.

\bibliographystyle{plainnat}
\bibliography{references}

\begin{thebibliography}{60}
\providecommand{\natexlab}[1]{#1}
\providecommand{\url}[1]{\texttt{#1}}
\expandafter\ifx\csname urlstyle\endcsname\relax
  \providecommand{\doi}[1]{doi: #1}\else
  \providecommand{\doi}{doi: \begingroup \urlstyle{rm}\Url}\fi

\bibitem[Achiam et~al.(2023)Achiam, Adler, Agarwal, Ahmad, Akkaya, Aleman, Almeida, Altenschmidt, Altman, Anadkat, et~al.]{achiam2023gpt}
Josh Achiam, Steven Adler, Sandhini Agarwal, Lama Ahmad, Ilge Akkaya, Florencia~Leoni Aleman, Diogo Almeida, Janko Altenschmidt, Sam Altman, Shyamal Anadkat, et~al.
\newblock Gpt-4 technical report.
\newblock \emph{arXiv preprint arXiv:2303.08774}, 2023.

\bibitem[Ahn et~al.(2022)Ahn, Brohan, Brown, Chebotar, Cortes, David, Finn, Fu, Gopalakrishnan, Hausman, et~al.]{ahn2022can}
Michael Ahn, Anthony Brohan, Noah Brown, Yevgen Chebotar, Omar Cortes, Byron David, Chelsea Finn, Chuyuan Fu, Keerthana Gopalakrishnan, Karol Hausman, et~al.
\newblock Do as i can, not as i say: Grounding language in robotic affordances.
\newblock \emph{arXiv preprint arXiv:2204.01691}, 2022.

\bibitem[Alayrac et~al.(2022)Alayrac, Donahue, Luc, Miech, Barr, Hasson, Lenc, Mensch, Millican, Reynolds, et~al.]{alayrac2022flamingo}
Jean-Baptiste Alayrac, Jeff Donahue, Pauline Luc, Antoine Miech, Iain Barr, Yana Hasson, Karel Lenc, Arthur Mensch, Katherine Millican, Malcolm Reynolds, et~al.
\newblock Flamingo: a visual language model for few-shot learning.
\newblock \emph{Advances in neural information processing systems}, 35:\penalty0 23716--23736, 2022.

\bibitem[Aldaco et~al.(2024)Aldaco, Armstrong, Baruch, Bingham, Chan, Draper, Dwibedi, Finn, Florence, Goodrich, et~al.]{aldaco2024aloha}
Jorge Aldaco, Travis Armstrong, Robert Baruch, Jeff Bingham, Sanky Chan, Kenneth Draper, Debidatta Dwibedi, Chelsea Finn, Pete Florence, Spencer Goodrich, et~al.
\newblock Aloha 2: An enhanced low-cost hardware for bimanual teleoperation.
\newblock \emph{arXiv preprint arXiv:2405.02292}, 2024.

\bibitem[Beyer et~al.(2024)Beyer, Steiner, Pinto, Kolesnikov, Wang, Salz, Neumann, Alabdulmohsin, Tschannen, Bugliarello, et~al.]{beyer2024paligemma}
Lucas Beyer, Andreas Steiner, Andr{\'e}~Susano Pinto, Alexander Kolesnikov, Xiao Wang, Daniel Salz, Maxim Neumann, Ibrahim Alabdulmohsin, Michael Tschannen, Emanuele Bugliarello, et~al.
\newblock Paligemma: A versatile 3b vlm for transfer.
\newblock \emph{arXiv preprint arXiv:2407.07726}, 2024.

\bibitem[Bharadhwaj et~al.(2024)Bharadhwaj, Vakil, Sharma, Gupta, Tulsiani, and Kumar]{bharadhwaj2023roboagentgeneralizationefficiencyrobot}
Homanga Bharadhwaj, Jay Vakil, Mohit Sharma, Abhinav Gupta, Shubham Tulsiani, and Vikash Kumar.
\newblock {RoboAgent}: Generalization and efficiency in robot manipulation via semantic augmentations and action chunking.
\newblock In \emph{2024 IEEE International Conference on Robotics and Automation (ICRA)}, pages 4788--4795. IEEE, 2024.

\bibitem[Brohan et~al.(2023)Brohan, Brown, Carbajal, Chebotar, Chen, Choromanski, Ding, Driess, Dubey, Finn, Florence, Fu, Arenas, Gopalakrishnan, Han, Hausman, Herzog, Hsu, Ichter, Irpan, Joshi, Julian, Kalashnikov, Kuang, Leal, Lee, Lee, Levine, Lu, Michalewski, Mordatch, Pertsch, Rao, Reymann, Ryoo, Salazar, Sanketi, Sermanet, Singh, Singh, Soricut, Tran, Vanhoucke, Vuong, Wahid, Welker, Wohlhart, Wu, Xia, Xiao, Xu, Xu, Yu, and Zitkovich]{brohan2023rt2visionlanguageactionmodelstransfer}
Anthony Brohan, Noah Brown, Justice Carbajal, Yevgen Chebotar, Xi~Chen, Krzysztof Choromanski, Tianli Ding, Danny Driess, Avinava Dubey, Chelsea Finn, Pete Florence, Chuyuan Fu, Montse~Gonzalez Arenas, Keerthana Gopalakrishnan, Kehang Han, Karol Hausman, Alexander Herzog, Jasmine Hsu, Brian Ichter, Alex Irpan, Nikhil Joshi, Ryan Julian, Dmitry Kalashnikov, Yuheng Kuang, Isabel Leal, Lisa Lee, Tsang-Wei~Edward Lee, Sergey Levine, Yao Lu, Henryk Michalewski, Igor Mordatch, Karl Pertsch, Kanishka Rao, Krista Reymann, Michael Ryoo, Grecia Salazar, Pannag Sanketi, Pierre Sermanet, Jaspiar Singh, Anikait Singh, Radu Soricut, Huong Tran, Vincent Vanhoucke, Quan Vuong, Ayzaan Wahid, Stefan Welker, Paul Wohlhart, Jialin Wu, Fei Xia, Ted Xiao, Peng Xu, Sichun Xu, Tianhe Yu, and Brianna Zitkovich.
\newblock Rt-2: Vision-language-action models transfer web knowledge to robotic control.
\newblock \emph{arXiv preprint arXiv:2307.15818}, 2023.

\bibitem[Cabi et~al.(2019)Cabi, Colmenarejo, Novikov, Konyushkova, Reed, Jeong, Zolna, Aytar, Budden, Vecerik, et~al.]{cabi2020scalingdatadrivenroboticsreward}
Serkan Cabi, Sergio~G{\'o}mez Colmenarejo, Alexander Novikov, Ksenia Konyushkova, Scott Reed, Rae Jeong, Konrad Zolna, Yusuf Aytar, David Budden, Mel Vecerik, et~al.
\newblock Scaling data-driven robotics with reward sketching and batch reinforcement learning.
\newblock \emph{arXiv preprint arXiv:1909.12200}, 2019.

\bibitem[Chi et~al.(2023)Chi, Xu, Feng, Cousineau, Du, Burchfiel, Tedrake, and Song]{chi2023diffusion}
Cheng Chi, Zhenjia Xu, Siyuan Feng, Eric Cousineau, Yilun Du, Benjamin Burchfiel, Russ Tedrake, and Shuran Song.
\newblock Diffusion policy: Visuomotor policy learning via action diffusion.
\newblock \emph{The International Journal of Robotics Research}, page 02783649241273668, 2023.

\bibitem[Collaboration et~al.(2023)Collaboration, Padalkar, Pooley, Jain, Bewley, Herzog, Irpan, Khazatsky, Rai, Singh, et~al.]{collaboration2023open}
OX-Embodiment Collaboration, A~Padalkar, A~Pooley, A~Jain, A~Bewley, A~Herzog, A~Irpan, A~Khazatsky, A~Rai, A~Singh, et~al.
\newblock {Open X-Embodiment}: Robotic learning datasets and {RT-X} models.
\newblock \emph{arXiv preprint arXiv:2310.08864}, 1\penalty0 (2), 2023.

\bibitem[Driess et~al.(2023)Driess, Xia, Sajjadi, Lynch, Chowdhery, Ichter, Wahid, Tompson, Vuong, Yu, et~al.]{driess2023palm}
Danny Driess, Fei Xia, Mehdi~SM Sajjadi, Corey Lynch, Aakanksha Chowdhery, Brian Ichter, Ayzaan Wahid, Jonathan Tompson, Quan Vuong, Tianhe Yu, et~al.
\newblock Palm-e: An embodied multimodal language model.
\newblock \emph{arXiv preprint arXiv:2303.03378}, 2023.

\bibitem[Du et~al.(2022)Du, Huang, Dai, Tong, Lepikhin, Xu, Krikun, Zhou, Yu, Firat, et~al.]{du2022glam}
Nan Du, Yanping Huang, Andrew~M Dai, Simon Tong, Dmitry Lepikhin, Yuanzhong Xu, Maxim Krikun, Yanqi Zhou, Adams~Wei Yu, Orhan Firat, et~al.
\newblock Glam: Efficient scaling of language models with mixture-of-experts.
\newblock In \emph{International Conference on Machine Learning}, pages 5547--5569. PMLR, 2022.

\bibitem[Ebert et~al.(2021)Ebert, Yang, Schmeckpeper, Bucher, Georgakis, Daniilidis, Finn, and Levine]{ebert2021bridgedataboostinggeneralization}
Frederik Ebert, Yanlai Yang, Karl Schmeckpeper, Bernadette Bucher, Georgios Georgakis, Kostas Daniilidis, Chelsea Finn, and Sergey Levine.
\newblock Bridge data: Boosting generalization of robotic skills with cross-domain datasets.
\newblock \emph{arXiv preprint arXiv:2109.13396}, 2021.

\bibitem[Esser et~al.(2024)Esser, Kulal, Blattmann, Entezari, M{\"u}ller, Saini, Levi, Lorenz, Sauer, Boesel, et~al.]{esser2024scaling}
Patrick Esser, Sumith Kulal, Andreas Blattmann, Rahim Entezari, Jonas M{\"u}ller, Harry Saini, Yam Levi, Dominik Lorenz, Axel Sauer, Frederic Boesel, et~al.
\newblock Scaling rectified flow transformers for high-resolution image synthesis.
\newblock In \emph{Forty-first International Conference on Machine Learning}, 2024.

\bibitem[Etukuru et~al.(2024)Etukuru, Naka, Hu, Lee, Mehu, Edsinger, Paxton, Chintala, Pinto, and Shafiullah]{etukuru2024robotutilitymodelsgeneral}
Haritheja Etukuru, Norihito Naka, Zijin Hu, Seungjae Lee, Julian Mehu, Aaron Edsinger, Chris Paxton, Soumith Chintala, Lerrel Pinto, and Nur Muhammad~Mahi Shafiullah.
\newblock Robot utility models: General policies for zero-shot deployment in new environments.
\newblock \emph{arXiv preprint arXiv:2409.05865}, 2024.

\bibitem[Fedus et~al.(2022)Fedus, Zoph, and Shazeer]{fedus2022switch}
William Fedus, Barret Zoph, and Noam Shazeer.
\newblock Switch transformers: Scaling to trillion parameter models with simple and efficient sparsity.
\newblock \emph{Journal of Machine Learning Research}, 23\penalty0 (120):\penalty0 1--39, 2022.

\bibitem[Fu et~al.(2024)Fu, Zhao, and Finn]{fu2024mobile}
Zipeng Fu, Tony~Z. Zhao, and Chelsea Finn.
\newblock Mobile aloha: Learning bimanual mobile manipulation with low-cost whole-body teleoperation.
\newblock In \emph{{Conference on Robot Learning (CoRL)}}, 2024.

\bibitem[Gupta et~al.(2018)Gupta, Murali, Gandhi, and Pinto]{gupta2018robotlearninghomesimproving}
Abhinav Gupta, Adithyavairavan Murali, Dhiraj~Prakashchand Gandhi, and Lerrel Pinto.
\newblock Robot learning in homes: Improving generalization and reducing dataset bias.
\newblock \emph{Advances in neural information processing systems}, 31, 2018.

\bibitem[He et~al.(2024)He, Fu, Liu, Wang, Xiao, Shu, Wang, Zhang, Yu, Li, et~al.]{he2024mars}
Wanggui He, Siming Fu, Mushui Liu, Xierui Wang, Wenyi Xiao, Fangxun Shu, Yi~Wang, Lei Zhang, Zhelun Yu, Haoyuan Li, et~al.
\newblock Mars: Mixture of auto-regressive models for fine-grained text-to-image synthesis.
\newblock \emph{arXiv preprint arXiv:2407.07614}, 2024.

\bibitem[Ho et~al.(2020)Ho, Jain, and Abbeel]{ho2020denoisingdiffusionprobabilisticmodels}
Jonathan Ho, Ajay Jain, and Pieter Abbeel.
\newblock Denoising diffusion probabilistic models.
\newblock \emph{Advances in neural information processing systems}, 33:\penalty0 6840--6851, 2020.

\bibitem[Jumper et~al.(2021)Jumper, Evans, Pritzel, Green, Figurnov, Ronneberger, Tunyasuvunakool, Bates, {\v{Z}}{\'\i}dek, Potapenko, et~al.]{jumper2021highly}
John Jumper, Richard Evans, Alexander Pritzel, Tim Green, Michael Figurnov, Olaf Ronneberger, Kathryn Tunyasuvunakool, Russ Bates, Augustin {\v{Z}}{\'\i}dek, Anna Potapenko, et~al.
\newblock Highly accurate protein structure prediction with alphafold.
\newblock \emph{Nature}, 596\penalty0 (7873):\penalty0 583--589, 2021.

\bibitem[Kalashnikov et~al.(2018)Kalashnikov, Irpan, Pastor, Ibarz, Herzog, Jang, Quillen, Holly, Kalakrishnan, Vanhoucke, et~al.]{kalashnikov2018qtoptscalabledeepreinforcement}
Dmitry Kalashnikov, Alex Irpan, Peter Pastor, Julian Ibarz, Alexander Herzog, Eric Jang, Deirdre Quillen, Ethan Holly, Mrinal Kalakrishnan, Vincent Vanhoucke, et~al.
\newblock Scalable deep reinforcement learning for vision-based robotic manipulation.
\newblock In \emph{Conference on robot learning}, pages 651--673. PMLR, 2018.

\bibitem[Khazatsky et~al.(2024)Khazatsky, Pertsch, Nair, Balakrishna, Dasari, Karamcheti, Nasiriany, Srirama, Chen, Ellis, et~al.]{khazatsky2024droid}
Alexander Khazatsky, Karl Pertsch, Suraj Nair, Ashwin Balakrishna, Sudeep Dasari, Siddharth Karamcheti, Soroush Nasiriany, Mohan~Kumar Srirama, Lawrence~Yunliang Chen, Kirsty Ellis, et~al.
\newblock {DROID}: A large-scale in-the-wild robot manipulation dataset.
\newblock \emph{arXiv preprint arXiv:2403.12945}, 2024.

\bibitem[Kim et~al.(2024)Kim, Pertsch, Karamcheti, Xiao, Balakrishna, Nair, Rafailov, Foster, Lam, Sanketi, et~al.]{kim2024openvla}
Moo~Jin Kim, Karl Pertsch, Siddharth Karamcheti, Ted Xiao, Ashwin Balakrishna, Suraj Nair, Rafael Rafailov, Ethan Foster, Grace Lam, Pannag Sanketi, et~al.
\newblock Openvla: An open-source vision-language-action model.
\newblock \emph{arXiv preprint arXiv:2406.09246}, 2024.

\bibitem[Lepikhin et~al.(2020)Lepikhin, Lee, Xu, Chen, Firat, Huang, Krikun, Shazeer, and Chen]{lepikhin2020gshard}
Dmitry Lepikhin, HyoukJoong Lee, Yuanzhong Xu, Dehao Chen, Orhan Firat, Yanping Huang, Maxim Krikun, Noam Shazeer, and Zhifeng Chen.
\newblock Gshard: Scaling giant models with conditional computation and automatic sharding.
\newblock \emph{arXiv preprint arXiv:2006.16668}, 2020.

\bibitem[Levine et~al.(2018)Levine, Pastor, Krizhevsky, Ibarz, and Quillen]{levine2016learninghandeyecoordinationrobotic}
Sergey Levine, Peter Pastor, Alex Krizhevsky, Julian Ibarz, and Deirdre Quillen.
\newblock Learning hand-eye coordination for robotic grasping with deep learning and large-scale data collection.
\newblock \emph{The International journal of robotics research}, 37\penalty0 (4-5):\penalty0 421--436, 2018.

\bibitem[Li et~al.(2022)Li, Choi, Chung, Kushman, Schrittwieser, Leblond, Eccles, Keeling, Gimeno, Lago, Hubert, Choy, de~Masson~d’Autume, Babuschkin, Chen, Huang, Welbl, Gowal, Cherepanov, Molloy, Mankowitz, Robson, Kohli, de~Freitas, Kavukcuoglu, and Vinyals]{li2022alphacode}
Yujia Li, David Choi, Junyoung Chung, Nate Kushman, Julian Schrittwieser, Rémi Leblond, Tom Eccles, James Keeling, Felix Gimeno, Agustin~Dal Lago, Thomas Hubert, Peter Choy, Cyprien de~Masson~d’Autume, Igor Babuschkin, Xinyun Chen, Po-Sen Huang, Johannes Welbl, Sven Gowal, Alexey Cherepanov, James Molloy, Daniel~J. Mankowitz, Esme~Sutherland Robson, Pushmeet Kohli, Nando de~Freitas, Koray Kavukcuoglu, and Oriol Vinyals.
\newblock Competition-level code generation with alphacode.
\newblock \emph{Science}, 378\penalty0 (6624):\penalty0 1092--1097, 2022.

\bibitem[Lipman et~al.(2022)Lipman, Chen, Ben-Hamu, Nickel, and Le]{lipman2022flow}
Yaron Lipman, Ricky~TQ Chen, Heli Ben-Hamu, Maximilian Nickel, and Matt Le.
\newblock Flow matching for generative modeling.
\newblock \emph{arXiv preprint arXiv:2210.02747}, 2022.

\bibitem[Liu et~al.(2024{\natexlab{a}})Liu, Akhgari, Visheratin, Kamko, Xu, Shrirao, Souza, Doshi, and Li]{liu2024playground}
Bingchen Liu, Ehsan Akhgari, Alexander Visheratin, Aleks Kamko, Linmiao Xu, Shivam Shrirao, Joao Souza, Suhail Doshi, and Daiqing Li.
\newblock Playground v3: Improving text-to-image alignment with deep-fusion large language models.
\newblock \emph{arXiv preprint arXiv:2409.10695}, 2024{\natexlab{a}}.

\bibitem[Liu et~al.(2024{\natexlab{b}})Liu, Li, Wu, and Lee]{liu2024visual}
Haotian Liu, Chunyuan Li, Qingyang Wu, and Yong~Jae Lee.
\newblock Visual instruction tuning.
\newblock \emph{Advances in neural information processing systems}, 36, 2024{\natexlab{b}}.

\bibitem[Liu et~al.(2024{\natexlab{c}})Liu, Orru, Vakil, Paxton, Shafiullah, and Pinto]{liu2024ok}
Peiqi Liu, Yaswanth Orru, Jay Vakil, Chris Paxton, Nur Muhammad~Mahi Shafiullah, and Lerrel Pinto.
\newblock Ok-robot: What really matters in integrating open-knowledge models for robotics.
\newblock \emph{arXiv preprint arXiv:2401.12202}, 2024{\natexlab{c}}.

\bibitem[Liu(2022)]{liu2022rectified}
Qiang Liu.
\newblock Rectified flow: A marginal preserving approach to optimal transport.
\newblock \emph{arXiv preprint arXiv:2209.14577}, 2022.

\bibitem[Mandlekar et~al.(2018)Mandlekar, Zhu, Garg, Booher, Spero, Tung, Gao, Emmons, Gupta, Orbay, et~al.]{mandlekar2018roboturkcrowdsourcingplatformrobotic}
Ajay Mandlekar, Yuke Zhu, Animesh Garg, Jonathan Booher, Max Spero, Albert Tung, Julian Gao, John Emmons, Anchit Gupta, Emre Orbay, et~al.
\newblock {RoboTurk}: A crowdsourcing platform for robotic skill learning through imitation.
\newblock In \emph{Conference on Robot Learning}, pages 879--893. PMLR, 2018.

\bibitem[Mandlekar et~al.(2023)Mandlekar, Nasiriany, Wen, Akinola, Narang, Fan, Zhu, and Fox]{mandlekar2023mimicgendatagenerationscalable}
Ajay Mandlekar, Soroush Nasiriany, Bowen Wen, Iretiayo Akinola, Yashraj Narang, Linxi Fan, Yuke Zhu, and Dieter Fox.
\newblock {MimicGen}: A data generation system for scalable robot learning using human demonstrations.
\newblock \emph{arXiv preprint arXiv:2310.17596}, 2023.

\bibitem[Ouyang et~al.(2022)Ouyang, Wu, Jiang, Almeida, Wainwright, Mishkin, Zhang, Agarwal, Slama, Ray, et~al.]{ouyang2022traininglanguagemodelsfollow}
Long Ouyang, Jeffrey Wu, Xu~Jiang, Diogo Almeida, Carroll Wainwright, Pamela Mishkin, Chong Zhang, Sandhini Agarwal, Katarina Slama, Alex Ray, et~al.
\newblock Training language models to follow instructions with human feedback.
\newblock \emph{Advances in neural information processing systems}, 35:\penalty0 27730--27744, 2022.

\bibitem[Peebles and Xie(2023)]{peebles2023scalable}
William Peebles and Saining Xie.
\newblock Scalable diffusion models with transformers.
\newblock In \emph{Proceedings of the IEEE/CVF International Conference on Computer Vision}, pages 4195--4205, 2023.

\bibitem[Pinto and Gupta(2016)]{pinto2015supersizingselfsupervisionlearninggrasp}
Lerrel Pinto and Abhinav Gupta.
\newblock Supersizing self-supervision: Learning to grasp from 50k tries and 700 robot hours.
\newblock In \emph{2016 IEEE international conference on robotics and automation (ICRA)}, pages 3406--3413. IEEE, 2016.

\bibitem[Polyak et~al.(2024)Polyak, Zohar, Brown, Tjandra, Sinha, Lee, Vyas, Shi, Ma, Chuang, et~al.]{polyak2024movie}
Adam Polyak, Amit Zohar, Andrew Brown, Andros Tjandra, Animesh Sinha, Ann Lee, Apoorv Vyas, Bowen Shi, Chih-Yao Ma, Ching-Yao Chuang, et~al.
\newblock Movie gen: A cast of media foundation models.
\newblock \emph{arXiv preprint arXiv:2410.13720}, 2024.

\bibitem[Radford et~al.(2021)Radford, Kim, Hallacy, Ramesh, Goh, Agarwal, Sastry, Askell, Mishkin, Clark, et~al.]{radford2021learning}
Alec Radford, Jong~Wook Kim, Chris Hallacy, Aditya Ramesh, Gabriel Goh, Sandhini Agarwal, Girish Sastry, Amanda Askell, Pamela Mishkin, Jack Clark, et~al.
\newblock Learning transferable visual models from natural language supervision.
\newblock In \emph{International conference on machine learning}, pages 8748--8763. PMLR, 2021.

\bibitem[Rombach et~al.(2022)Rombach, Blattmann, Lorenz, Esser, and Ommer]{rombach2022high}
Robin Rombach, Andreas Blattmann, Dominik Lorenz, Patrick Esser, and Bj{\"o}rn Ommer.
\newblock High-resolution image synthesis with latent diffusion models.
\newblock In \emph{Proceedings of the IEEE/CVF conference on computer vision and pattern recognition}, pages 10684--10695, 2022.

\bibitem[Saharia et~al.(2022)Saharia, Chan, Saxena, Li, Whang, Denton, Ghasemipour, Gontijo~Lopes, Karagol~Ayan, Salimans, et~al.]{saharia2022photorealistic}
Chitwan Saharia, William Chan, Saurabh Saxena, Lala Li, Jay Whang, Emily~L Denton, Kamyar Ghasemipour, Raphael Gontijo~Lopes, Burcu Karagol~Ayan, Tim Salimans, et~al.
\newblock Photorealistic text-to-image diffusion models with deep language understanding.
\newblock \emph{Advances in neural information processing systems}, 35:\penalty0 36479--36494, 2022.

\bibitem[Sanh(2019)]{sanh2019distilbert}
V~Sanh.
\newblock Distilbert, a distilled version of bert: Smaller, faster, cheaper and lighter.
\newblock \emph{arXiv preprint arXiv:1910.01108}, 2019.

\bibitem[Shafiullah et~al.(2023)Shafiullah, Rai, Etukuru, Liu, Misra, Chintala, and Pinto]{shafiullah2023bringingrobotshome}
Nur Muhammad~Mahi Shafiullah, Anant Rai, Haritheja Etukuru, Yiqian Liu, Ishan Misra, Soumith Chintala, and Lerrel Pinto.
\newblock On bringing robots home.
\newblock \emph{arXiv preprint arXiv:2311.16098}, 2023.

\bibitem[Shazeer(2019)]{shazeer2019fast}
Noam Shazeer.
\newblock Fast transformer decoding: One write-head is all you need.
\newblock \emph{arXiv preprint arXiv:1911.02150}, 2019.

\bibitem[Shazeer et~al.(2017)Shazeer, Mirhoseini, Maziarz, Davis, Le, Hinton, and Dean]{shazeer2017outrageously}
Noam Shazeer, Azalia Mirhoseini, Krzysztof Maziarz, Andy Davis, Quoc Le, Geoffrey Hinton, and Jeff Dean.
\newblock Outrageously large neural networks: The sparsely-gated mixture-of-experts layer.
\newblock \emph{arXiv preprint arXiv:1701.06538}, 2017.

\bibitem[Sohl-Dickstein et~al.(2015)Sohl-Dickstein, Weiss, Maheswaranathan, and Ganguli]{sohldickstein2015deepunsupervisedlearningusing}
Jascha Sohl-Dickstein, Eric Weiss, Niru Maheswaranathan, and Surya Ganguli.
\newblock Deep unsupervised learning using nonequilibrium thermodynamics.
\newblock In \emph{International conference on machine learning}, pages 2256--2265. PMLR, 2015.

\bibitem[Steiner et~al.(2021)Steiner, Kolesnikov, Zhai, Wightman, Uszkoreit, and Beyer]{steiner2022trainvitdataaugmentation}
Andreas Steiner, Alexander Kolesnikov, Xiaohua Zhai, Ross Wightman, Jakob Uszkoreit, and Lucas Beyer.
\newblock How to train your vit? data, augmentation, and regularization in vision transformers.
\newblock \emph{arXiv preprint arXiv:2106.10270}, 2021.

\bibitem[Team et~al.(2023)Team, Anil, Borgeaud, Alayrac, Yu, Soricut, Schalkwyk, Dai, Hauth, Millican, et~al.]{team2023gemini}
Gemini Team, Rohan Anil, Sebastian Borgeaud, Jean-Baptiste Alayrac, Jiahui Yu, Radu Soricut, Johan Schalkwyk, Andrew~M Dai, Anja Hauth, Katie Millican, et~al.
\newblock Gemini: a family of highly capable multimodal models.
\newblock \emph{arXiv preprint arXiv:2312.11805}, 2023.

\bibitem[Team et~al.(2024{\natexlab{a}})Team, Mesnard, Hardin, Dadashi, Bhupatiraju, Pathak, Sifre, Rivi{\`e}re, Kale, Love, et~al.]{team2024gemma}
Gemma Team, Thomas Mesnard, Cassidy Hardin, Robert Dadashi, Surya Bhupatiraju, Shreya Pathak, Laurent Sifre, Morgane Rivi{\`e}re, Mihir~Sanjay Kale, Juliette Love, et~al.
\newblock Gemma: Open models based on gemini research and technology.
\newblock \emph{arXiv preprint arXiv:2403.08295}, 2024{\natexlab{a}}.

\bibitem[Team et~al.(2024{\natexlab{b}})Team, Ghosh, Walke, Pertsch, Black, Mees, Dasari, Hejna, Kreiman, Xu, et~al.]{team2024octo}
Octo~Model Team, Dibya Ghosh, Homer Walke, Karl Pertsch, Kevin Black, Oier Mees, Sudeep Dasari, Joey Hejna, Tobias Kreiman, Charles Xu, et~al.
\newblock Octo: An open-source generalist robot policy.
\newblock \emph{arXiv preprint arXiv:2405.12213}, 2024{\natexlab{b}}.

\bibitem[Vaswani et~al.(2017)Vaswani, Shazeer, Parmar, Uszkoreit, Jones, Gomez, Kaiser, and Polosukhin]{vaswani2017attention}
Ashish Vaswani, Noam Shazeer, Niki Parmar, Jakob Uszkoreit, Llion Jones, Aidan~N Gomez, \L~ukasz Kaiser, and Illia Polosukhin.
\newblock Attention is all you need.
\newblock In \emph{Advances in Neural Information Processing Systems}, volume~30, 2017.

\bibitem[Walke et~al.(2023)Walke, Black, Zhao, Vuong, Zheng, Hansen-Estruch, He, Myers, Kim, Du, et~al.]{walke2023bridgedata}
Homer~Rich Walke, Kevin Black, Tony~Z Zhao, Quan Vuong, Chongyi Zheng, Philippe Hansen-Estruch, Andre~Wang He, Vivek Myers, Moo~Jin Kim, Max Du, et~al.
\newblock {BridgeData} v2: A dataset for robot learning at scale.
\newblock In \emph{Conference on Robot Learning}, pages 1723--1736. PMLR, 2023.

\bibitem[Wei et~al.(2021)Wei, Bosma, Zhao, Guu, Yu, Lester, Du, Dai, and Le]{wei2021finetuned}
Jason Wei, Maarten Bosma, Vincent~Y Zhao, Kelvin Guu, Adams~Wei Yu, Brian Lester, Nan Du, Andrew~M Dai, and Quoc~V Le.
\newblock Finetuned language models are zero-shot learners.
\newblock \emph{arXiv preprint arXiv:2109.01652}, 2021.

\bibitem[Wei et~al.(2022)Wei, Tay, Bommasani, Raffel, Zoph, Borgeaud, Yogatama, Bosma, Zhou, Metzler, et~al.]{wei2022emergent}
Jason Wei, Yi~Tay, Rishi Bommasani, Colin Raffel, Barret Zoph, Sebastian Borgeaud, Dani Yogatama, Maarten Bosma, Denny Zhou, Donald Metzler, et~al.
\newblock Emergent abilities of large language models.
\newblock \emph{arXiv preprint arXiv:2206.07682}, 2022.

\bibitem[Wen et~al.(2024)Wen, Zhu, Li, Zhu, Wu, Xu, Liu, Cheng, Shen, Peng, Feng, and Tang]{wen2024tinyvlafastdataefficientvisionlanguageaction}
Junjie Wen, Yichen Zhu, Jinming Li, Minjie Zhu, Kun Wu, Zhiyuan Xu, Ning Liu, Ran Cheng, Chaomin Shen, Yaxin Peng, Feifei Feng, and Jian Tang.
\newblock Tinyvla: Towards fast, data-efficient vision-language-action models for robotic manipulation.
\newblock \emph{arXiv preprint arXiv:2409.12514}, 2024.

\bibitem[Yu et~al.(2016)Yu, Bauza, Fazeli, and Rodriguez]{yu2016more}
Kuan-Ting Yu, Maria Bauza, Nima Fazeli, and Alberto Rodriguez.
\newblock More than a million ways to be pushed. a high-fidelity experimental dataset of planar pushing.
\newblock In \emph{2016 IEEE/RSJ international conference on intelligent robots and systems (IROS)}, pages 30--37. IEEE, 2016.

\bibitem[Zhao et~al.(2023)Zhao, Kumar, Levine, and Finn]{zhao2023learning}
Tony~Z Zhao, Vikash Kumar, Sergey Levine, and Chelsea Finn.
\newblock Learning fine-grained bimanual manipulation with low-cost hardware.
\newblock \emph{arXiv preprint arXiv:2304.13705}, 2023.

\bibitem[Zhao et~al.(2024)Zhao, Tompson, Driess, Florence, Ghasemipour, Finn, and Wahid]{zhao2024alohaunleashedsimplerecipe}
Tony~Z Zhao, Jonathan Tompson, Danny Driess, Pete Florence, Kamyar Ghasemipour, Chelsea Finn, and Ayzaan Wahid.
\newblock Aloha unleashed: A simple recipe for robot dexterity.
\newblock \emph{arXiv preprint arXiv:2410.13126}, 2024.

\bibitem[Zhou et~al.(2024)Zhou, Yu, Babu, Tirumala, Yasunaga, Shamis, Kahn, Ma, Zettlemoyer, and Levy]{zhou2024transfusion}
Chunting Zhou, Lili Yu, Arun Babu, Kushal Tirumala, Michihiro Yasunaga, Leonid Shamis, Jacob Kahn, Xuezhe Ma, Luke Zettlemoyer, and Omer Levy.
\newblock Transfusion: Predict the next token and diffuse images with one multi-modal model.
\newblock \emph{arXiv preprint arXiv:2408.11039}, 2024.

\bibitem[Zhu et~al.(2024)Zhu, Zhu, Li, Wen, Xu, Liu, Cheng, Shen, Peng, Feng, et~al.]{zhu2024scalingdiffusionpolicytransformer}
Minjie Zhu, Yichen Zhu, Jinming Li, Junjie Wen, Zhiyuan Xu, Ning Liu, Ran Cheng, Chaomin Shen, Yaxin Peng, Feifei Feng, et~al.
\newblock Scaling diffusion policy in transformer to 1 billion parameters for robotic manipulation.
\newblock \emph{arXiv preprint arXiv:2409.14411}, 2024.

\end{thebibliography}

\appendix

\section{}

\subsection{Contributions}
\label{app:contributions}

The authors contributed to the following areas (listed alphabetically):

\noindent \textbf{Data and operations:} Noah Brown, Michael Equi, Chelsea Finn, Niccolo Fusai, Lachy Groom, Liyiming Ke, Suraj Nair, Lucy Shi, and Anna Walling.

\noindent \textbf{Evaluation experiments:} Kevin Black, Michael Equi, Chelsea Finn, Brian Ichter, Liyiming Ke, Adrian Li-Bell, Suraj Nair, Karl Pertsch, and Lucy Shi.

\noindent \textbf{Model design:} Kevin Black, Brian Ichter, Sergey Levine, Karl Pertsch, Lucy Shi, and Quan Vuong.

\noindent \textbf{Post-training:} Michael Equi, Chelsea Finn, Liyiming Ke, Adrian Li-Bell, Suraj Nair, and Lucy Shi.

\noindent \textbf{Pre-training:} Kevin Black, Danny Driess, Brian Ichter, Sergey Levine, Karl Pertsch, Lucy Shi, and Quan Vuong.

\noindent \textbf{Robot hardware:} Noah Brown, Adnan Esmail, Chelsea Finn, Tim Jones, and Mohith Mothukuri.

\noindent \textbf{Robot software:} Karol Hausman, Szymon Jakubczak, Sergey Levine, James Tanner, and Haohuan Wang.

\noindent \textbf{Training infrastructure:} Kevin Black, Michael Equi, Sergey Levine, Adrian Li-Bell, Suraj Nair, Quan Vuong, Haohuan Wang, and Ury Zhilinsky.

\noindent \textbf{Writing and illustration:} Kevin Black, Chelsea Finn, Lachy Groom, Karol Hausman, Brian Ichter, Sergey Levine, and Quan Vuong.

\subsection{Model Architecture Details}
\label{app:architecture}

In this section, we provide a full description of the model architecture. We follow the PaliGemma VLM~\citep{beyer2024paligemma} design, with the following differences: (1) additional input and output projections for the robotics-specific tokens, including the state vector $\bq_t$ and action vectors $\bA_t = [\ba_t, ..., \ba_{t+H-1}]$, (2) an additional MLP for incorporating the flow matching timestep information $\tau$, and (3) a second, smaller set of weights for the action expert.

\textbf{Additional inputs and outputs.} The standard PaliGemma architecture takes in a sequence of images $[\bI^1_t, ..., \bI^n_t]$ followed by a language prompt $\ell_t$. We add an input $\bq_t$ for the robot's proprioceptive state, which is mapped to the transformer embedding dimension using a linear projection. The final set of input tokens correspond to the noisy action chunk \mbox{$\bA^\tau_t = [\ba^\tau_t, ..., \ba^\tau_{t+H-1}]$}, with the number of tokens equal to the action horizon ($H = 50$ for our tasks). We only use the transformer outputs corresponding to the $H$ noisy actions, which are decoded into $\bv_\theta(\bA_t^\tau, \bo_t)$ using a linear projection.

\textbf{Incorporating the flow matching timestep.} The noisy action chunk $\bA^\tau_t$ is mapped to the transformer's embedding dimension using an MLP that also incorporates the flow matching timestep $\tau$. For each noisy action $\ba^\tau_{t'}$, the expression for the corresponding embedding that is fed into the transformer is \mbox{$W_3 \cdot \text{swish} (W_2 \cdot \text{concat} (W_1 \cdot \ba^\tau_{t'}, \phi(\tau)))$}, where $\phi : \mathbb{R} \to \mathbb{R}^w$ is a sinusoidal positional encoding function~\citep{vaswani2017attention}, $W_1 \in \mathbb{R}^{w \times d}$, $W_2 \in \mathbb{R}^{w \times 2w}$, $W_3 \in \mathbb{R}^{w \times w}$, $d$ is the action dimension, and $w$ is the embedding dimension (or \textit{width}) of the action expert.

\textbf{Attention mask.} \ModelSymbol{} uses a blockwise causal attention mask with 3 blocks: $[\bI^1_t, ..., \bI^n_t, \ell_t]$, $[\bq_t]$, and $[\ba^\tau_t, ..., \ba^\tau_{t+H-1}]$. Within each block, there is full bidirectional attention, whereas the tokens in each block cannot attend to the tokens in future blocks. The first block includes the input modalities from PaliGemma's VLM pre-training, which are prevented from attending to future blocks (which include new inputs) to minimize distribution shift from said pre-training. The robot state $\bq_t$ is its own block because it does not change with each flow matching integration step; preventing it from attending to the final block allows its corresponding keys and values to be cached during sampling. The final block corresponds to the noisy actions $\bA^\tau_t$, which can attend to the full input sequence.

\textbf{Action expert.} \ModelSymbol{} is implemented as a single transformer with two sets of weights (also known as experts~\citep{shazeer2017outrageously}), where each token is routed to one of the experts; the weights interact only through the transformer's self-attention layers. The images and language prompt, $[\bI^1_t, ..., \bI^n_t, \ell_t]$, are routed to the larger VLM backbone, which we initialize from PaliGemma.  The inputs not seen during VLM pre-training, $[\bq_t, \bA^\tau_t]$, are routed to the action expert. PaliGemma is based on the Gemma 2B~\citep{team2024gemma} language model, which uses multi-query attention~\citep{shazeer2019fast} and a configuration of \{\textit{width}=2048, \textit{depth}=18, \textit{mlp\_dim}=16,384, \textit{num\_heads}=18, \textit{num\_kv\_heads}=1, \textit{head\_dim}=256\}. Since the experts interact only in the self-attention layers, \textit{width} and \textit{mlp\_dim} do not necessarily need to match between experts. To speed up inference (which requires multiple forward passes of the action expert), we downsize the action expert to \{\textit{width}=1024, \textit{mlp\_dim}=4096\}, resulting in a parameter count of $\sim$300M.

\begin{figure}
    \centering
    \includegraphics[width=0.75\linewidth]{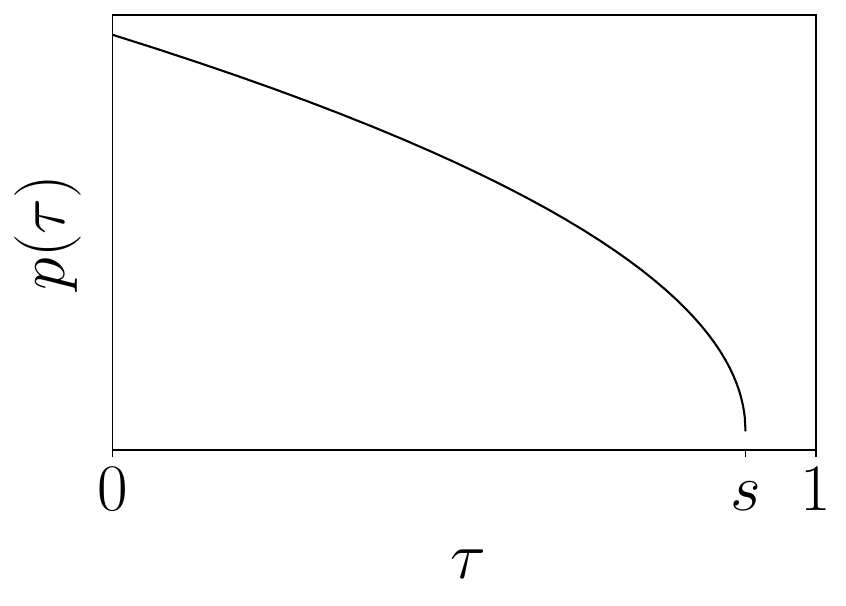}
    \caption{\textbf{Flow matching timestep sampling distribution}. We sample $\tau$ from a shifted beta distribution that emphasizes lower timesteps (corresponding to noisier actions), and does not sample timesteps at all above a cutoff value $s$. We use $s = 0.999$ in our experiments.}
    \label{fig:timestep-sampling}
\end{figure}

\textbf{Sampling the flow matching timestep.} The original flow matching papers~\citep{lipman2022flow,liu2022rectified} sample the flow matching timestep from a uniform distribution: $\tau \sim \mathcal{U}(0, 1)$. \citet{esser2024scaling} instead propose sampling from a logit-normal distribution that emphasizes the middle timesteps; the authors posit that at high timesteps (low noise levels), the model needs only to learn the identity function, and at low timesteps (high noise levels), the model needs only to learn the mean of the data distribution. However, we hypothesize that the task of action prediction is subtly different from high-resolution image synthesis --- while it may be relatively easy to predict the mean image conditioned on a text label, predicting the mean action conditioned on a robot observation (i.e., learning $\E [\bA_t | \bo_t]$) is a much harder problem; this is because the observation $\bo_t$ is very \textit{informative} in that it should constrain the distribution of possible actions much more than a text label constrains the distribution of possible images. As a result, we designed a timestep sampling distribution that emphasizes low timesteps (high noise levels); additionally, timesteps above a given threshold $s$ are not sampled at all, since they are not needed so long as the integration step $\delta$ is greater than $1-s$. The distribution is given by $p(\tau) = \text{Beta}(\frac{s-\tau}{s}; 1.5, 1)$ and is visualized in Figure~\ref{fig:timestep-sampling}. We use $s = 0.999$ in our experiments, which allows for $\delta > \frac{1}{1000}$, or up to 1,000 integration steps.

\subsection{Non-VLM Baseline Architecture}
\label{app:small_model}

Our baseline architecture \ModelSymbol-small is \emph{not} based on a VLM backbone. Hence, we use it to evaluate the benefits of VLM-pre-training.
We design it to be sufficiently expressive to fit our large dataset while still providing good performance when trained from scratch. This model has about 470M parameters, and differs from our main model in the following ways: (1) We use DistilBERT~\citep{sanh2019distilbert} to encode the language tokens of the language command $\ell_t$, since this model does not use a language model backbone; (2) The action expert cross-attends to the outputs of the observation encoder, akin to a traditional encoder-decoder transformer~\citep{vaswani2017attention}, rather than our main model which is more like a decoder-only mixture of experts~\citep{shazeer2017outrageously}; (3) The images are encoded with a smaller pre-trained ViT encoder (specifically, the R26-S-32 ResNet-ViT hybrid from \citet{steiner2022trainvitdataaugmentation}); (4) The ViT image encoders do not share weights; (5) The transformer backbone that encodes the observations (which comes after the ViT image encoders) is not pre-trained on Internet data; (6) The action expert uses the DiT architecture~\citep{peebles2023scalable} rather than the Gemma architecture, and hence incorporates the flow-matching timestep $\tau$ using AdaLN-Zero layers. Besides this, the models are broadly similar: both use pre-trained ViT image encoders, both use separate weights for the observation encoder and the action expert, both take in the same observation format, and both perform 10 steps of flow matching to predict the action chunk.

\subsection{Inference}
\label{app:inference}

Recall that our model takes an observation $\bo_t = [\bI^1_t , ... , \bI^n_t, \ell_t, \bq_t]$ and the noisy actions $\bA_t^\tau$ and outputs the vector field that needs to be integrated to obtain the next flow matching step, $\bv_t^\tau$. Each time we predict a new action chunk $\bA_t$, we must encode each of the images $\bI^1_t , ... , \bI^n_t$, run a forward pass on the tokens corresponding to $\bo_t$, and then run 10 steps of flow matching, where each step requires running a forward pass on the tokens corresponding to $\bA_t^\tau$ (the keys and values corresponding to $\bo_t$ are cached). Table \ref{tab:inference_time} summarizes the computation time for this operation with 3 camera images. The operations were timed on an NVIDIA GeForce RTX 4090 consumer-grade GPU. For the mobile robot, inference was done off-board over a Wi-Fi connection, adding a small amount of network latency. Further optimizations, quantization, and other improvements might further reduce inference times.

Since the model generates an entire $H$-step action chunk at once, we can execute up to $H$ actions before we need to run inference again. However, we may run inference more often than that, as well as combine actions from different inference calls using various aggregation strategies. We tried temporal ensembling~\citep{zhao2023learning} early on and found that it hurt policy performance, so we opted not to aggregate actions and instead execute action chunks open-loop. For the 20Hz UR5e and Franka robots, we run inference every 0.8 seconds (after executing 16 actions), and for all other robots, which run at 50Hz, we run inference every 0.5 seconds (after executing 25 actions).
\begin{table}[h]
\centering
\begin{tabular}{rl}
\toprule
\textbf{model part} & \textbf{inference time} \\
\midrule
image encoders & 14 ms  \\
observation forward pass & 32 ms \\
x10 action forward pass (flow) & 27 ms \\
network latency (if off-board) & 13 ms \\
\midrule
total on-board inference & 73 ms \\
total off-board inference & 86 ms \\
\bottomrule
\end{tabular}
\caption{Inference time of our model on an NVIDIA GeForce RTX 4090 GPU.}
\label{tab:inference_time}
\end{table}

\subsection{Evaluation Details}
\label{app:score}

For each task, we design a score rubric that measures progress on the task, and use this for our quantitative results. We describe this rubric for each task below:

\noindent \textbf{A. Evaluating the base model} 

\noindent \textbf{Shirt folding:} Shirt folding is recorded as either success or failure. We begin each shirt folding eval by laying the shirt flat on the table. Success is defined as having folded in the sleeves and performed one half-fold along the length of the shirt. Our eval includes 4 small t-shirts and 1 medium t-shirt. We run 2 evals for each item for a maximum of 15000 steps or approximately 5 minutes each.

\noindent \textbf{Bussing easy:} This task is scored out of 7, where there are 7 different objects on the table, and 1 point is given for each correctly sorted object.

\noindent \textbf{Bussing hard:} This task is scored out of 12, where there are 12 different objects on the table, and 1 point is given for each correctly sorted object. This version of the task includes particularly challenging settings, like a chopstick on top of a piece of trash. 

\noindent \textbf{Grocery bagging:} This task is scored out of 7. For each 7 grocery items, a point is given for putting it in the bag. 

\noindent \textbf{Toast out of toaster:} This task is scored out of 4. For each piece of toast, 1 point is given for picking it from the toaster and another for putting it on the plate. 

\noindent \textbf{B. Language instruction following.} The policy is scored on successfully repositioning each object and whether it follows instructions.

\noindent \textbf{Bussing:} The robot has to follow the command to pick up the correct object and place each of them into the correct receptacle. The robot receives 12 objects in total and around 30 instructions in one episode.

\noindent \textbf{Table setting:} The robot arranges all dishes, utensils, and napkins and makes adjustments according to language specification. The robot receives 7 objects in total and around 20 instructions in one episode.

\noindent \textbf{Grocery bagging:} The robot picks up the correct item (among bag of coffee beans, bag of barley, bag of marshmallow, cat food, spaghetti, bag of seaweed, bag of almonds), and bags them into a paper bag. The robot receives 7 objects in total and around 14 instructions in one episode.

\noindent \textbf{C. Learning new dexterous tasks}

\noindent \textbf{Stack bowls:} This task is scored out of 3. One point for each of two bowls stacked in larger bowls, and one for the neatness of the final product.

\noindent \textbf{Towel folding:} This task is scored out of 3. One point for the first half-fold of the towel, one point for the second half-fold of the towel, and one point for neatness of the final product.

\noindent \textbf{Tupperware in microwave:} This task is scored out of 4. One point for opening the microwave, one point for picking up the Tupperware, one point for putting the Tupperware in the microwave, and one point for closing the microwave.

\noindent \textbf{Paper towel replacement:} This task is scored out of 4. One point is given for grasping the old roll, and another point is given for removing it. Then, one point is given for grasping the new paper towel roll, and the final point is given for placing it on the dispenser. 

\noindent \textbf{Items in drawer:} This task is scored out of 5. One point for opening the drawer, one point for each of 3 items picked and placed into the drawer, and one point for closing the drawer. 

\noindent \textbf{D. Mastering complex multi-stage tasks}

\noindent \textbf{Laundry folding:} This task is scored out of 4. Our evaluation includes five items, three shirts of size M, L, and XL and two shorts of size 28 and 36. We perform two trials for each item, and the items left to be evaluated start randomly crumpled in a laundry bin (while previously evaluated items start in a folded stack). One point is given for picking an item out of the bin and putting it on the table. Another point is given for flattening the shirt or shorts. A third point is granted for folding the shirt or shorts. A final point is given for either placing the item in the corner of the table (if it is the first item evaluated), or stacking it onto an existing stack of folded clothes. We run each eval for a maximum of 15000 steps or approximately 5 minutes.

\noindent \textbf{Mobile laundry:} This evaluation follows the same protocol as laundry folding. The three shirts are sized M, M, and XL, and the shorts are sized 32 and 31 W.

\noindent \textbf{Table bussing:} This task is scored out of 12, where there are 12 different objects on the table, and 1 point is given for each correctly sorted object. This version of the task includes particularly challenging settings, like a chopstick on top of a piece of trash. 

\noindent \textbf{Box building:} This task is scored out of 5. One point is given for successfully picking up the box to begin the task. One point is given for folding the box in half, so the flaps can be closed. One point is given for closing the right flap. One point is given for closing the left flap. The final point is given for neatly centering the final product.

\noindent \textbf{Packing eggs:} This task is scored out of 7. One point for each egg placed in the correct slot in the carton, and one point for closing the lid.

\noindent \textbf{Packing food:} This task is scored out of 5. One point for picking up the plate of food, one point for each of 3 food items placed in the to-go box, and one point for closing the to-go box.

\noindent \textbf{Dryer unloading:} This task involves having the robot approach a dyer with a laundry basket and unload the clothes into the basket. We score this eval out of five, where one point is given for properly approaching the dryer. Another for placing the laundry basket on the stool. A third for opening the dryer. A fourth for putting all the clothes in the basket and a fifth point for closing the dryer. We eval with 3 shirts and 2 shorts that start in a random configuration inside the dryer.

\end{document}